\documentclass[journal]{IEEEtran}
\usepackage{cite}
\usepackage{amsmath,amssymb,amsfonts}
\usepackage{algorithmic}
\usepackage{graphicx}
\usepackage{textcomp}
\usepackage{lineno,hyperref}
\usepackage{times}
\usepackage{epsfig}
\usepackage{graphicx}
\usepackage{amsmath}
\usepackage{amssymb}
\usepackage{caption}

\usepackage{booktabs}
\usepackage{epstopdf}

\usepackage{subeqnarray}
\usepackage{color}
\usepackage{amstext}
\usepackage{verbatim}
\usepackage{array}
\usepackage{boldline}
\usepackage{makecell}

\usepackage{subfigure}
\usepackage{setspace}
 \usepackage{threeparttable}

\usepackage{chngpage}
\usepackage{multirow}
\graphicspath{{figure/},{figure1/},{figure2/},{figure3/}}

\begin{document}
%
\title{Multi-level Wavelet Convolutional Neural Networks}
%
%
%

\author{Pengju Liu,
        Hongzhi Zhang,
        Wei Lian,
        and~Wangmeng Zuo
\thanks{P. Liu is with School of Computer Science and Technology, Harbin Institute of Technology, China,
 e-mail: lpj008@126.com.}
\thanks{H. Zhang and W. Zuo are with Harbin Institute of Technology.}
\thanks{W. Lian is with Department of Computer Science, Changzhi University, China.}%
}

\maketitle

\begin{abstract}
In computer vision, convolutional networks (CNNs) often adopts pooling to enlarge   receptive field which has the advantage of low computational complexity.
However, pooling can cause information loss and thus is detrimental to further operations such as features extraction and analysis.
Recently, dilated filter has been proposed to trade off between receptive field size and efficiency. 
But the accompanying gridding effect can cause a sparse sampling of input images with checkerboard patterns.
To address this problem, in this paper, we propose a novel multi-level wavelet CNN (MWCNN) model to achieve better trade-off between receptive field size and computational efficiency.
The core idea is to embed wavelet transform into CNN architecture to reduce the resolution of feature maps while at the same time, increasing receptive field. 
Specifically, MWCNN for image restoration is based on U-Net architecture, and inverse wavelet transform (IWT) is deployed to reconstruct the high resolution (HR) feature maps. 
The proposed MWCNN can also be viewed as an improvement of dilated filter and a generalization of average pooling, and can be applied to not only image restoration tasks, but also any CNNs requiring a pooling operation.
The experimental results demonstrate effectiveness of the proposed MWCNN for tasks such as  image denoising, single image super-resolution, JPEG image artifacts removal and object classification.
The code and pre-trained models will be given at \url{https://github.com/lpj-github-io/MWCNNv2}.
\end{abstract}


%
\IEEEpeerreviewmaketitle

\section{Introduction}
Nowadays,  convolutional networks have become the  dominant technique behind many computer vision tasks, \textit{e.g.} image restoration~\cite{dong2016image,Kim2015Accurate,Zhang2016Beyond,lai2017deep,Shi2016Real} and object classification~\cite{krizhevsky2012imagenet,simonyan2014very,yu2015multi,he2016deep,he2016identity}.
With continual progress, CNNs are extensively and easily learned on large-scale datasets, speeded up by increasingly advanced GPU devices, and often achieve state-of-the-art performance in comparison with traditional methods.
The reason that CNN is popular in computer vision can be contributed to two aspects.
First, existing CNN-based solutions dominate on several simple tasks by outperforming other methods with a large margin, such as single image super-resolution (SISR)~\cite{dong2016image,Kim2015Accurate,Ledig2017Photo}, image denoising~\cite{Zhang2016Beyond}, image deblurring~\cite{zhang2017learning}, compressed imaging~\cite{adler2016deep}, and object classification~\cite{krizhevsky2012imagenet}.
Second, CNNs can be treated as a modular part and plugged into traditional method, which also promotes the widespread use of CNNs~\cite{zhang2017learning,romano2016little,yan2015image}.

\begin{figure}[!t]
\vspace{-0ex}
\begin{center}
  \vspace{-0.0ex}
  \includegraphics[width=0.42\textwidth]{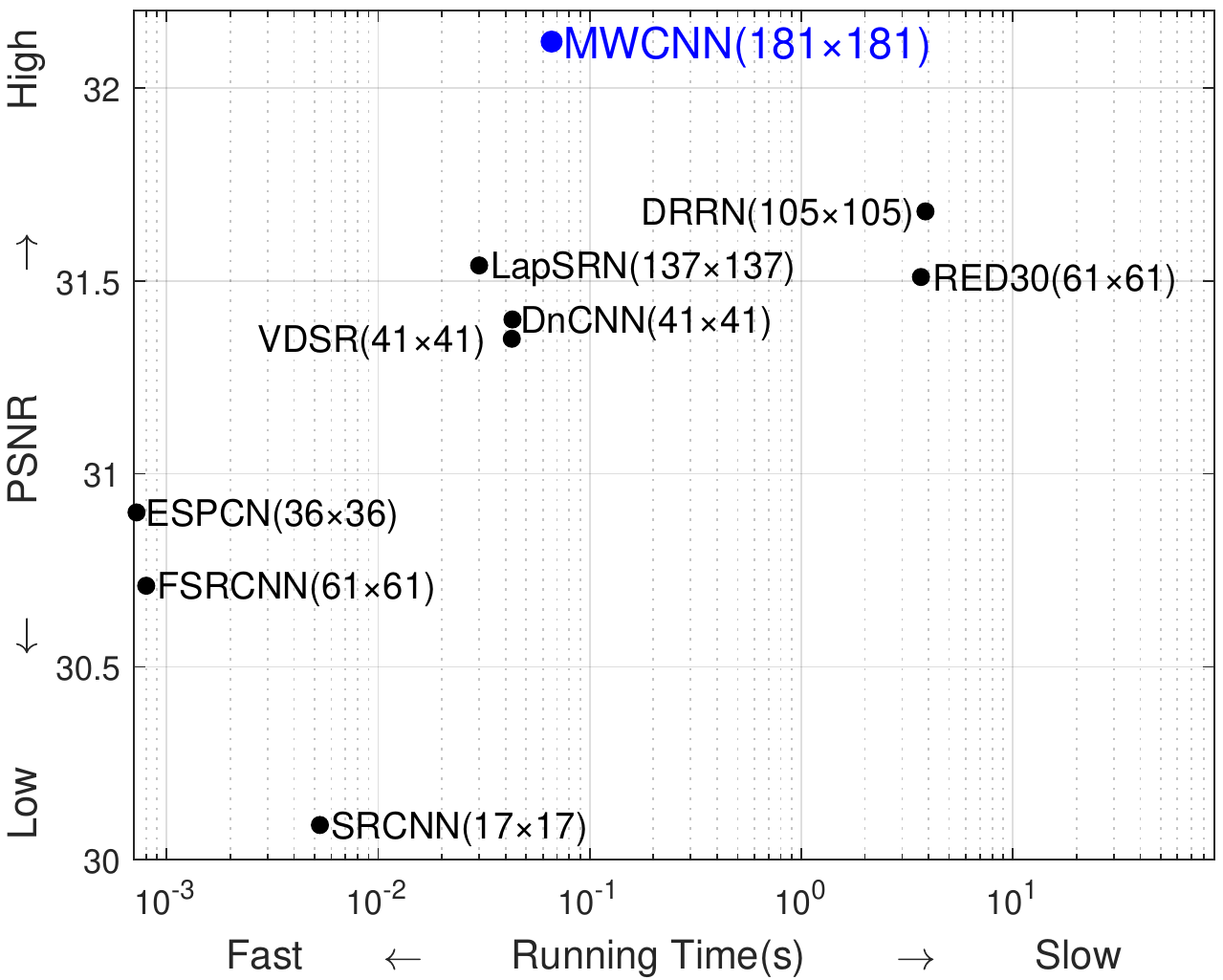}
  \vspace{-2.0ex}
\end{center}
  \caption{The running time vs. PSNR value of representative CNN models, including SRCNN~\cite{dong2016image}, FSRCNN~\cite{Dong2016Accelerating}, ESPCN~\cite{Shi2016Real}, VDSR~\cite{Kim2015Accurate}, DnCNN~\cite{Zhang2016Beyond}, RED30~\cite{Mao2016Image}, LapSRN~\cite{lai2017deep}, DRRN~\cite{Tai2017Image}, MemNet~\cite{tai2017memnet} and our MWCNN. The receptive field of each model are also provided. The PSNR and time are evaluated on Set5 with the scale factor $\times4$ running on a GTX1080 GPU.}
  \vspace{-0ex}
  \label{fig:receptive}
\end{figure}

Actually, CNNs in computer vision can be viewed as a non-linear map from the input image to the target. 
In general, larger receptive field is helpful for improving fitting ability of CNNs and promoting accurate performance by taking more spatial context into account. 
Generally, the receptive field can be enlarged by either increasing the network depth, enlarging filter size or using pooling operation.
But increasing the network depth or enlarging filter size can inevitably result in higher computational cost.
Pooling can enlarge receptive field and guarantee efficiency by directly reducing spatial resolution of feature map.
Nevertheless, it may result in information loss.
Recently, dilated filtering~\cite{yu2015multi} is proposed to trade off between receptive field size and efficiency by inserting ``zero holes'' in convolutional filtering.
However, the receptive field of dilated filtering with fixed factor greater than 1 only takes into account a sparse sampling of the input with checkerboard patterns, thus it can lead to inherent suffering from gridding effect~\cite{Wang2017Understanding}.
Based on the above analysis, one can see that we should be careful when enlarging receptive field if we want to avoid both increasing computational burden and incurring the potential performance sacrifice.
As can be seen from Figure~\ref{fig:receptive}, even though  DRRN~\cite{Tai2017Image} and MemNet~\cite{tai2017memnet} enjoy larger receptive fields and higher PSNR performances than VDSR~\cite{Kim2015Accurate} and DnCNN~\cite{Zhang2016Beyond}, their speed nevertheless are orders of magnitude slower.

In an attempt to address the problems stated previously, we propose an efficient CNN based approach aiming at trading off between performance and efficiency.
More specifically, we propose a multi-level wavelet CNN (MWCNN) by utilizing discrete wavelet transform (DWT) to replace the pooling operations.
Due to invertibility of DWT, none of image information or intermediate features are lost by the proposed downsampling scheme.
Moreover, both frequency and location information of feature maps are captured by DWT~\cite{daubechies1990wavelet,daubechies1992ten}, which is helpful for preserving detailed texture when using multi-frequency feature representation.
More specifically, we adopt inverse wavelet transform (IWT) with expansion convolutional layer to restore resolutions of feature maps in image restoration tasks, where U-Net architecture~\cite{Ronneberger2015U} is used as a backbone network architecture.
Also, element-wise summation is adopted to combine feature maps, thus enriching feature representation.

In terms of relation with relevant works, we show that dilated filtering can be interpreted as a special variant of MWCNN, and the proposed method is more general and effective in enlarging receptive field.
Using an ensemble of such networks trained with embedded multi-level wavelet, we achieve PSNR/SSIM value that improves upon the best known results in image restoration tasks such as image denoising, SISR and JPEG image artifacts removal.
For the task of object classification, the proposed MWCNN can achieve higher performance than when adopting pooling layers.
As shown in Figure~\ref{fig:receptive}, although MWCNN is moderately slower than LapSRN~\cite{lai2017deep}, DnCNN~\cite{Zhang2016Beyond} and VDSR~\cite{Kim2015Accurate},
MWCNN can have a much larger receptive field and achieve higher PSNR value.

This paper is an extension of our previous work~\cite{liu2018multi}.
Compared to the former work~\cite{liu2018multi}, we propose a more general approach for improving performance, further extend it to high-level task and provide more analysis and discussions.
To sum up, the contributions of this work include:
\begin {itemize}
   \item A novel MWCNN model to enlarge receptive field with better tradeoff between efficiency and restoration performance by introducing wavelet transform.
   \item Promising detail preserving due to the good time-frequency localization property of DWT.
   \item A general approach to embedding wavelet transform in any CNNs where pooling operation is employed.
   \item State-of-the-art performance on image denoising, SISR, JPEG image artifacts removal, and classification.
 \end {itemize}

The remainder of the paper is organized as follows.
Sec.~\ref{sec:related} briefly reviews the development of CNNs for image restoration and classification. 
Sec.~\ref{sec:method} describes the proposed MWCNN model in detail.
Sec.~\ref{sec:exp} reports the experimental results in terms of performance evaluation.
Finally, Sec.~\ref{sec:con} concludes the paper.

\section{Related Work}\label{sec:related}
In this section, the development of CNNs for image restoration tasks is briefly reviewed. In particular, we discuss  relevant works on incorporating DWT in CNNs.
Finally, relevant  object classification works are introduced.

\subsection{Image Restoration}
Image restoration aims at recovering the latent clean image $\mathbf{x}$ from its degraded observation $\mathbf{y}$.
For decades, researches on image restoration have been done from the view points of both prior modeling and discriminative learning~\cite{Banham1977Digital,Chen2015Trainable, dabov2007image, gu2014weighted, schmidt2014shrinkage, wright2009robust}.
Recently, with the booming development, CNNs based methods achieve state-of-the-art performance over the traditional methods.

\subsubsection{Improving Performance and Efficiency of CNNs for Image Restoration}
In the early attempt, the CNN-based methods don't work so well on some image restoration tasks.
For example, the methods of \cite{Agostinelli2013Robust,jain2009natural,Xie2012Image} could not achieve state-of-the-art denoising performance compared to BM3D~\cite{dabov2007image} in 2007.
In \cite{burger2012image}, multi-layer perception (MLP) achieved comparable performance as BM3D by learning the mapping from noise patches to clean patches.
In 2014, Dong \textit{et al.}~\cite{dong2016image} for the first time adopted only a 3-layer FCN without pooling for SISR, which realizes only a small receptive field but achieves state-of-the-art performance.
Then, Dong \textit{et al.}~\cite{Dong2016Compression} proposed a 4-layer ARCNN for JPEG image artifacts reduction.

Recently, deeper networks are increasingly  used for image restoration.
For SISR, Kim \textit{et al.}~\cite{Kim2015Accurate} stacked a 20-layer CNN with residual learning and adjustable gradient clipping.
Subsequently, some works, for example, very deep network~\cite{lim2017enhanced,Zhang2016Beyond,Zhang2018Rcan}, symmetric skip connections~\cite{Mao2016Image}, residual units~\cite{Ledig2017Photo}, Laplacian pyramid~\cite{lai2017deep}, and recursive architecture~\cite{kim2016deeply,Tai2017Image}, had also been suggested to enlarge receptive field.
However, the receptive field of those methods is enlarged with the increase of  network depth, which may has limited potential to extend to deeper network.

For better tradeoff between speed and performance, a 7-layer FCN with dilated filtering was presented as a denoiser by Zhang \textit{et al.}~\cite{zhang2017learning}.
Santhanam \textit{et al.}~\cite{santhanam2017generalized} adopt pooling/unpooling to obtain and aggregate multi-context representation for image denoising.
In \cite{zhang2018ffdnet}, Zhang~\textit{et al.} considered to operate the CNN denoiser  on downsampled subimages .
Guo \textit{et al.}~\cite{guo2019toward} utilized U-Net~\cite{Ronneberger2015U} based CNN as non-blind denoiser.
On account of the speciality of SISR,  the receptive field size and efficiency could be better traded off by taking the low-resolution (LR) images as input and zooming in on features with upsampling operation~\cite{Dong2016Accelerating,Shi2016Real,johnson2016perceptual}.
Nevertheless, this strategy can only be adopted for SISR, and are not suitable for other tasks, such as  image denoising and JPEG image artifacts removal.

\subsubsection{Universality of Image Restoration}
On account of the similarity of tasks such as image denoising, SISR, and JPEG image artifacts removal, the model suggested for one task may be easily extended to other image restoration tasks simply by retraining the same network.
For example, both DnCNN~\cite{Zhang2016Beyond} and MemNet~\cite{tai2017memnet} had been evaluated on all these three tasks.
Moreover, CNN denoisers can also serve as a kind of plug-and-play prior.
Thus, any restoration tasks can be tackled by sequentially applying the CNN denoisers via incorporating with unrolled inference~\cite{zhang2017learning}.
To provide an explicit functional for defining regularization induced by denoisers,  Romano \textit{et al.}~\cite{romano2016little} further proposed a regularization-by-denoising framework.
In \cite{riegler2015conditioned} and \cite{Zhang_2018_CVPR}, LR image with blur kernel is incorporated into CNNs for non-blind SR.
These methods not only promote the application of CNN in low level vision, but also present solutions to deploying CNN denoisers for other image restoration tasks.

\subsubsection{Incorporating DWT in CNNs}
Several studies have also been given to incorporate wavelet transform into CNN. 
Bae \textit{et al.}~\cite{bae2017beyond} proposed a wavelet residual network (WavResNet) with the discovery that CNN learning can benefit from learning on wavelet subbands with features having more channels.
For recovering missing details in subbands, Guo \textit{et al.}~\cite{guo2017deep} proposed a deep wavelet super-resolution (DWSR) method.
Subsequently, deep convolutional framelets (DCF)~\cite{Han2017Framing, Ye2017Deep} had been developed for low-dose CT and inverse problems. 
However, only one-level wavelet decomposition is considered in WavResNet and DWSR which may restrict the application of wavelet transform.
Inspired by the view point of decomposition, DCF independently processes each subband, which spontaneously ignores the dependency between these subbands.
 In contrast, multi-level wavelet transform is considered by our MWCNN to enlarge receptive field where computational burden is barely increased. 

\subsection{Object Classification}
The AlexNet~\cite{krizhevsky2012imagenet} is a 8-layers network for object classification, and for the first time achieved state-of-the-art performance than other methods on the ILSVRC2012 dataset.
In this method, different sized filters are adopted for extracting and enhancing features.
However, Simonyan \textit{et al.}~\cite{simonyan2014very} found that using only $3\times3$ sized convolutional filter with deeper architecture can realize larger receptive field and achieve better performance than AlexNet.
Yu \textit{et al.}~\cite{yu2015multi} adopted the dilated convolution to enlarge the receptive field size without increasing the computation burden.
Later, residual block~\cite{he2016deep,he2016identity}, inception model~\cite{szegedy2015going}, pyramid architecture~\cite{han2017deep},  doubly CNN~\cite{zhai2016doubly} and other architectures~\cite{zagoruyko2016wide,huang2017densely} were proposed for object classification.
Some measures on pooling operation, such as parallel grid pooling  \cite{Takeki18Parallel} and gated mixture of second-order pooling \cite{wang2018global}, were also proposed to enhance feature extractor or feature representation to promote performance.
In general, pooling operation, such as average pooling and max pooling, is often adopted for downsampling features and enlarging receptive field, but it can result in significant information loss.
To avoid this downside, we adopt DWT as our downsampling layer by replacing pooling operation without changing the main architecture, resulting in more power for enhancing feature representation.

\section{Method}\label{sec:method}
In this section, we first briefly introduce the concept of multi-level wavelet packet transform (WPT) and provide our motivation.
We then formally present our MWCNN based on multi-level WPT, and describe its network architecture for image restoration and object classification.
Finally, discussion is presented to analyze the connection of MWCNN with average pooling and dilated filtering.

\begin{figure}[!t]
\begin{center}
\vspace{-0.0ex}
\hspace{-0ex}
\subfigure[Multi-level WPT architecture]{
\begin{minipage}[c]{0.48\textwidth}
\centering
  \includegraphics[width=0.98\linewidth]{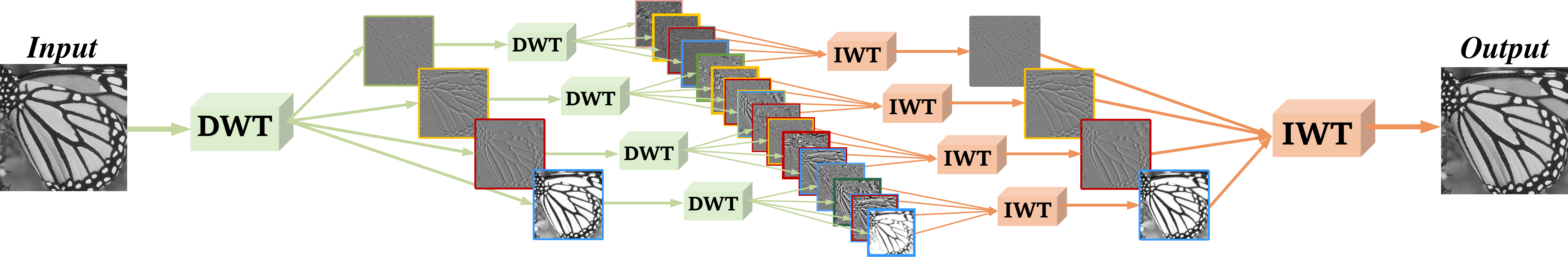}
  \label{fig:wpt_arch}
\end{minipage}%
\vspace{-2.0ex}
}
\subfigure[Embed CNN blocks]{
\begin{minipage}[c]{0.48\textwidth}
\centering
  \includegraphics[width=0.98\linewidth]{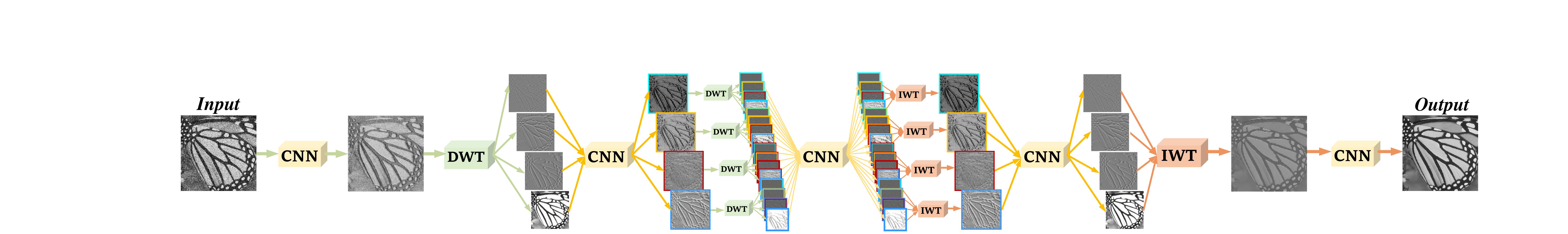}
  \label{fig:mwcnn_arch_g}
\end{minipage}%
}
\vspace{-0ex}
\end{center}
  \caption{From WPT to MWCNN.
  Intuitively, WPT can be seen as a special case of our MWCNN without CNN blocks as shown in (a) and (b).
  By inserting CNN blocks to WPT, we design our MWCNN as (b).
  Obviously, our MWCNN is a generalization of multi-level WPT, and reduces to WPT when each CNN block becomes the identity mapping. 
  }
 \vspace{-0ex}

 \label{fig:wpt}
\end{figure}

\begin{figure*}[!t]
\begin{center}
\vspace{-0.0ex}
\hspace{-0ex}
\subfigure{
\begin{minipage}[c]{0.93\textwidth}
\centering
  \includegraphics[width=0.98\linewidth]{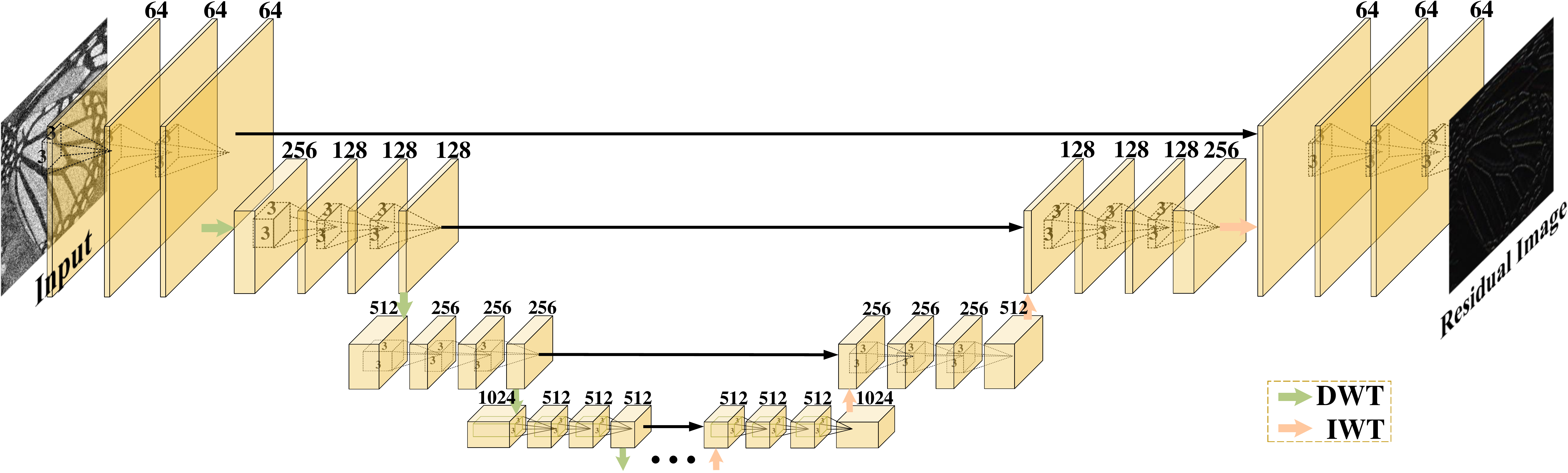}
  \label{fig:final_arch2}
\end{minipage}%
}
\vspace{-3.0ex}
\end{center}
  \caption{Multi-level wavelet-CNN architecture.
  It consists of two parts: the contracting and expanding subnetworks.
  Each solid box corresponds to a multi-channel feature map.
  And the number of channels is annotated on the top of the boxes.
  The number of convolutional layers is set to 24.
  Moreover, our MWCNN can be further extended to higher level (\textit{e.g.}, $\geq 4$) by duplicating the configuration of the 3rd level subnetwork. 
  }
 \vspace{-0ex}

 \label{fig:Architecture}
\end{figure*}

\subsection{From multi-level WPT to MWCNN}\label{sec:wpt2mw}

Given an image $\mathbf{x}$, we can use 2D DWT~\cite{mallat1989theory} with four convolutional filters,  \textit{i.e.} low-pass filter $\mathbf{f}_{LL}$, and high-pass filters $\mathbf{f}_{LH}$, $\mathbf{f}_{HL}$, $\mathbf{f}_{HH}$, to decompose $\mathbf{x}$ into four subband images, \textit{i.e.} $\mathbf{x}_{LL}$, $\mathbf{x}_{LH}$, $\mathbf{x}_{HL}$, and $\mathbf{x}_{HH}$.
Note that the four filters have  fixed parameters with convolutional stride 2 during the transformation.
Taking Haar wavelet as an example, four filters are defined as
\begin{equation}\label{eq:haar_ll}
\!\!
  \mathbf{f}_{LL} \!\!=\!\! \begin{bmatrix}
                    1 \!&\! 1\\
                    1 \!&\! 1
                    \end{bmatrix},
    \mathbf{f}_{LH} \!\!=\!\! \begin{bmatrix}
                    -1 \!\!\!&\!\!\! -1\\
                    1 \!\!\!&\!\!\! 1
                    \end{bmatrix},
    \mathbf{f}_{HL} \!\!=\!\! \begin{bmatrix}
                    -1 \!\!&\!\! 1\\
                    -1 \!\!&\!\! 1
                    \end{bmatrix},
      \mathbf{f}_{HH} \!\!=\!\! \begin{bmatrix}
                    1 \!\!\!\!&\!\!\! -1\\
                    -1 \!\!\!\!&\!\!\! 1
                    \end{bmatrix}.
\end{equation}

It is evident that $\mathbf{f}_{LL}$, $\mathbf{f}_{LH}$, $\mathbf{f}_{HL}$, and $\mathbf{f}_{HH}$ are orthogonal to each other and form a $4\times4$ invertible matrix.
The operation of DWT is defined as
$\mathbf{x}_{LL} \!\!=\!\! (\mathbf{f}_{LL} \!\otimes\! \mathbf{x})\!\!\downarrow_2$, $\mathbf{x}_{LH} \!\!=\!\! (\mathbf{f}_{LH} \! \otimes \! \mathbf{x})\!\! \downarrow_2$, $\mathbf{x}_{HL} \!\!=\!\! (\mathbf{f}_{HL} \! \otimes \! \mathbf{x}) \!\!\downarrow_2$, and $\mathbf{x}_{HH} \!\!=\!\! (\mathbf{f}_{HH} \! \otimes \! \mathbf{x})\!\! \downarrow_2$, where $\otimes$ denotes convolution operator, and $\downarrow_2$ means the standard downsampling operator with factor 2.
In other words, DWT mathematically involves four fixed convolution filters  with stride 2  to implement downsampling operator.
Moreover, according to the theory of Haar transform~\cite{mallat1989theory}, the $(i,j)$-th value of $\mathbf{x}_{LL}$, $\mathbf{x}_{LH}$, $\mathbf{x}_{HL}$ and $\mathbf{x}_{HH}$ after 2D Haar transform can be written as
\begin{equation}
\!\!\!\!\!\!
\begin{cases}
\!\!\mathbf{x}_{LL}(i,j) \!\!=\!\!  \mathbf{x}(2i\!\!-\!\!1,2j\!\!-\!\!1) \!\!+\!\! \mathbf{x}(2i\!\!-\!\!1,2j)  \!\!+\!\! \mathbf{x}(2i,2j\!\!-\!\!1) \!\!+\!\! \mathbf{x}(2i,2j)\\
\!\!\mathbf{x}_{LH}(i,j) \!\!=\!\!  -\mathbf{x}(2i\!\!-\!\!1,2j\!\!-\!\!1) \!\!-\!\! \mathbf{x}(2i\!\!-\!\!1,2j)  \!\!+\!\! \mathbf{x}(2i,2j\!\!-\!\!1) \!\!+\!\! \mathbf{x}(2i,2j)\\
\!\!\mathbf{x}_{HL}(i,j) \!\!=\!\!  -\mathbf{x}(2i\!\!-\!\!1,2j\!\!-\!\!1) \!\!+\!\! \mathbf{x}(2i\!\!-\!\!1,2j) \!\!-\!\! \mathbf{x}(2i,2j\!\!-\!\!1) \!\!+\!\! \mathbf{x}(2i,2j)\\
\!\!\mathbf{x}_{HH}(i,j) \!\!=\!\!  \mathbf{x}(2i\!\!-\!\!1,2j\!\!-\!\!1) \!\!-\!\! \mathbf{x}(2i\!\!-\!\!1,2j)  \!\!-\!\! \mathbf{x}(2i,2j\!\!-\!\!1) \!\!+\!\! \mathbf{x}(2i,2j).
\end{cases}
\label{eq:Haar_trans}
\end{equation}

Although the downsampling operation is deployed, due to the biorthogonal property of DWT, the original image $\mathbf{x}$ can be accurately reconstructed without information loss by the IWT, \textit{i.e.}, $\mathbf{x} = IWT(\mathbf{x}_{LL}, \mathbf{x}_{LH}, \mathbf{x}_{HL}, \mathbf{x}_{HH})$.
For the Haar wavelet, the IWT can defined as following:
\begin{equation}
\!\!\!\!\!\!
\begin{cases}
\!\!\mathbf{x}(2i\!\!-\!\!1,2j\!\!-\!\!1)  \!\!=\!\!   \left(\mathbf{x}_{LL}(i,j) \!\!-\!\! \mathbf{x}_{LH}(i,j) \!\!-\!\! \mathbf{x}_{HL}(i,j) \!\!+\!\!\mathbf{x}_{HH}(i,j)\right)/4, \\
\!\!\mathbf{x}(2i\!\!-\!\!1,2j) \quad  \!\!=\!\!   \left(\mathbf{x}_{LL}(i,j) \!\!-\!\! \mathbf{x}_{LH}(i,j) \!\!+\!\! \mathbf{x}_{HL}(i,j) \!\!-\!\!\mathbf{x}_{HH}(i,j)\right)/4, \\
\!\!\mathbf{x}(2i\!\!-\!\!1,2j) \quad  \!\!=\!\!   \left(\mathbf{x}_{LL}(i,j) \!\!+\!\! \mathbf{x}_{LH}(i,j) \!\!- \!\!\mathbf{x}_{HL}(i,j)\!\! -\!\!\mathbf{x}_{HH}(i,j)\right)/4, \\
\!\!\mathbf{x}(2i,2j) \quad\quad  \!\!=\!\!    \left(\mathbf{x}_{LL}(i,j) \!\!+\!\! \mathbf{x}_{LH}(i,j) \!\!+\!\! \mathbf{x}_{HL}(i,j) \!\!+\!\! \mathbf{x}_{HH}(i,j)\right)/4.
\end{cases}
\label{eq:iHaar_trans}
\end{equation}

Generally, the subband images $\mathbf{x}_{LL}$, $\mathbf{x}_{LH}$, $\mathbf{x}_{HL}$, and $\mathbf{x}_{HH}$ can be sequentially decomposed by DWT for further processing in multi-level WPT~\cite{akansu2001multiresolution,daubechies1992ten}.
To get results of two-level WPT, DWT is separately utilized to decompose each subband image $\mathbf{x}_{i}$ ($i =$ $LL$, $LH$, $HL$, or $HH$) into four subband images $\mathbf{x}_{i,LL}$, $\mathbf{x}_{i,LH}$, $\mathbf{x}_{i,HL}$, and $\mathbf{x}_{i,LL}$.
Recursively, the results of three or higher levels WPT can be obtained.
Correspondingly, the reconstruction of each level subband images are implemented by completely inverse operation via IWT.
The above-mentioned process of decomposition and reconstruction of an image are illustrated in Figure \ref{fig:wpt_arch}.
If we treat the filers of WPT  as convolutional filters with pre-defined weights, one can see that WPT is a special case of FCN without the nonlinearity layers.
Obviously, the original image $\mathbf{x}$ can be first decomposed by WPT and then accurately reconstructed by inverse WPT without any information loss.

In image processing applications such as image denoising and compression, some operations, \textit{e.g.}, soft-threshold and quantization, are usually required to process the decomposition part~\cite{chang2000adaptive,lewis1992image} as shown in Figure.~\ref{fig:wpt_arch}.
These operations can be treated as some kind of nonlinearity tailored to specific task.
In this work, we further extend WPT to multi-level wavelet-CNN (MWCNN) by plugging CNN blocks into traditional WPT-based method as illustrated in Figure \ref{fig:mwcnn_arch_g}.
Due to the biorthogonal property of WPT, our MWCNN can use subsampling and upsampling operations safely without incurring information loss.
Obviously, our MWCNN is a generalization of multi-level WPT, and reduces to WPT when each CNN block becomes the identity mapping.
Moreover, DWT can be treated as downsampling operation and extend to any CNNs where pooling operation is required.

\begin{figure*}[!htbp]
\scriptsize{
\vspace{-0ex}
\begin{center}
\subfigure[]{
\begin{minipage}[c]{0.315\textwidth}
\centering
  \includegraphics[width=0.89\linewidth]{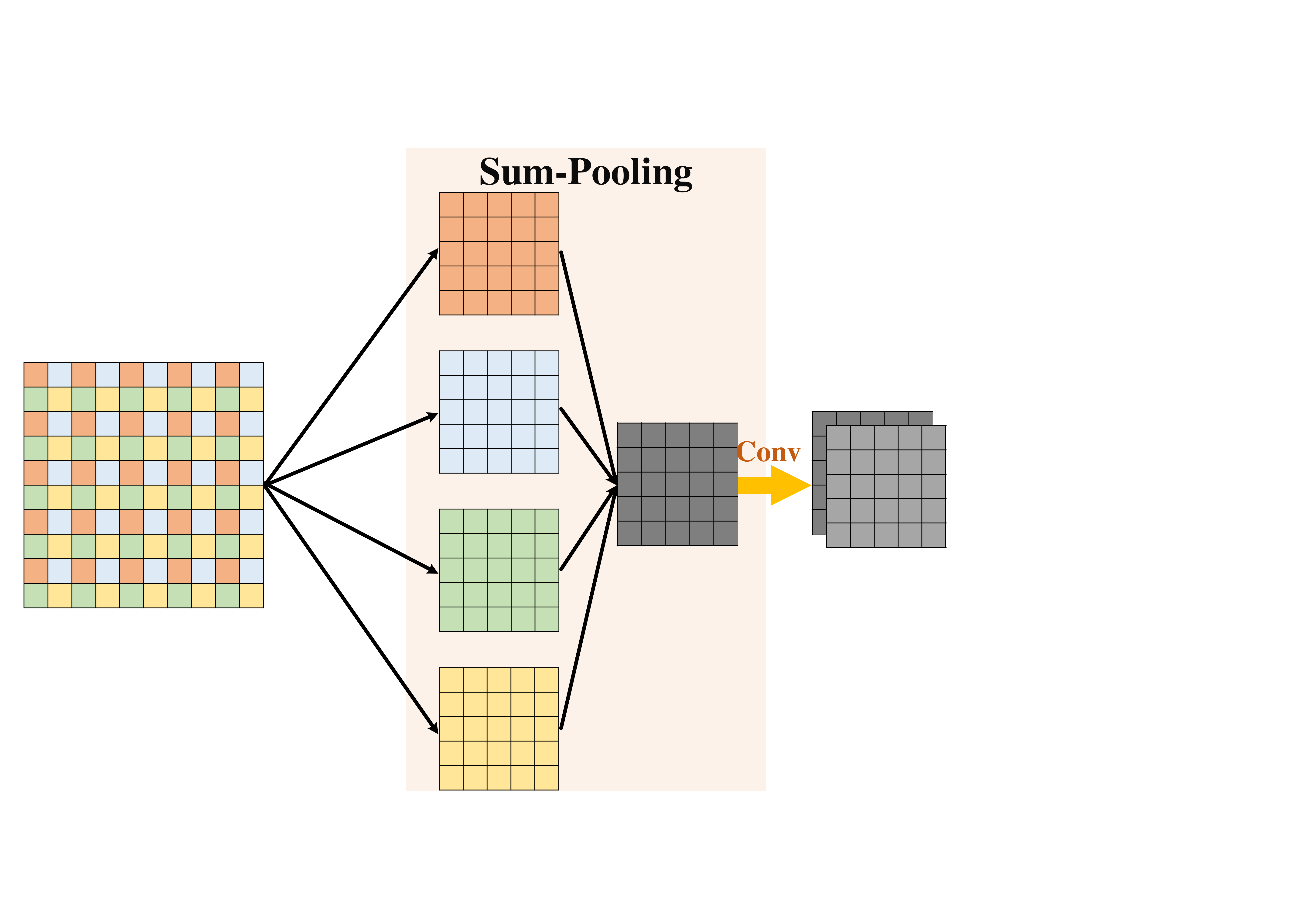}
  \label{fig:con2pooling}
\end{minipage}%
}
\subfigure[]{
\begin{minipage}[c]{0.315\textwidth}
\centering
  \includegraphics[width=0.99\linewidth]{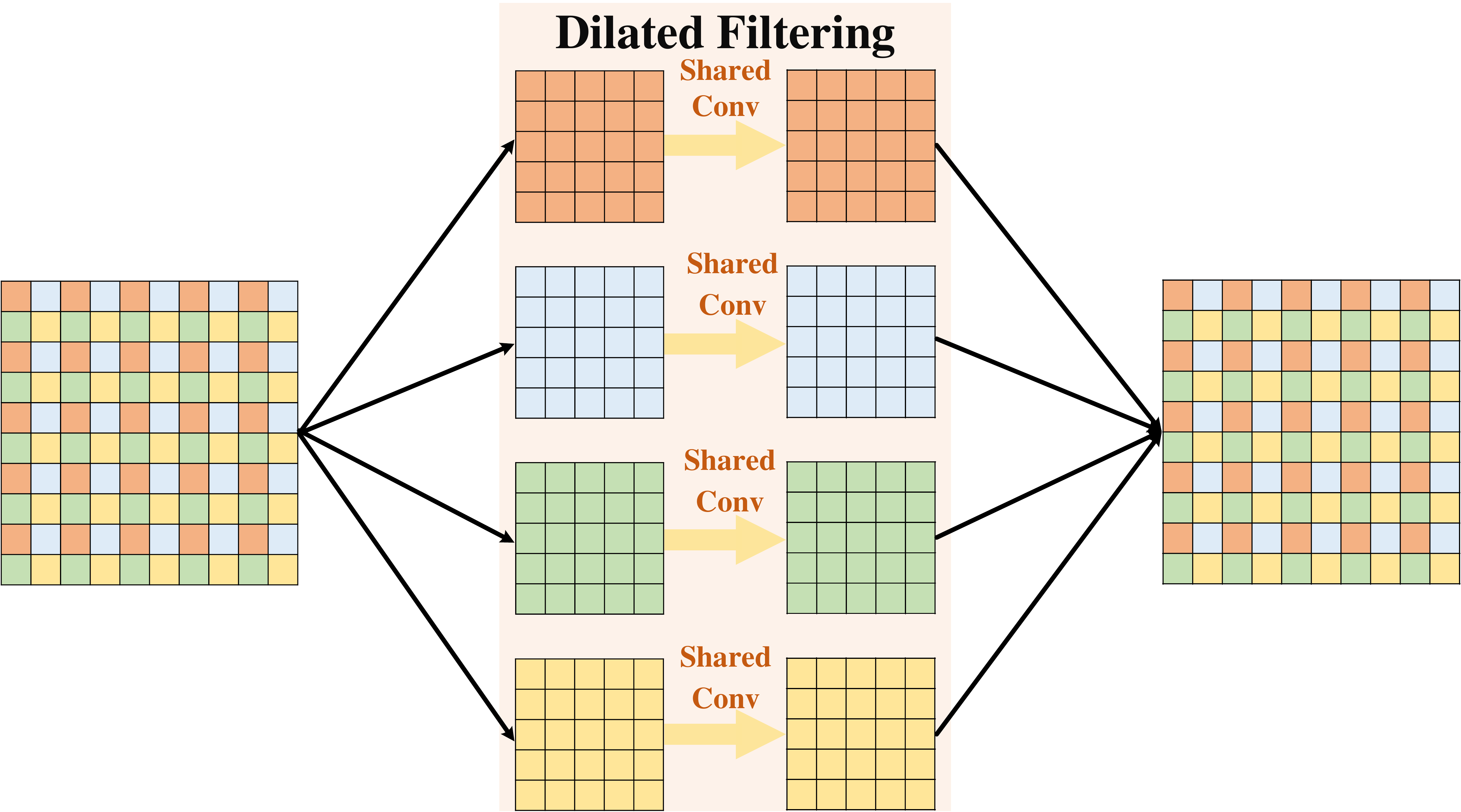}
  \label{fig:dailated}
\end{minipage}%
}
\subfigure[]{
\begin{minipage}[c]{0.33\textwidth}
\centering
  \includegraphics[width=0.99\linewidth]{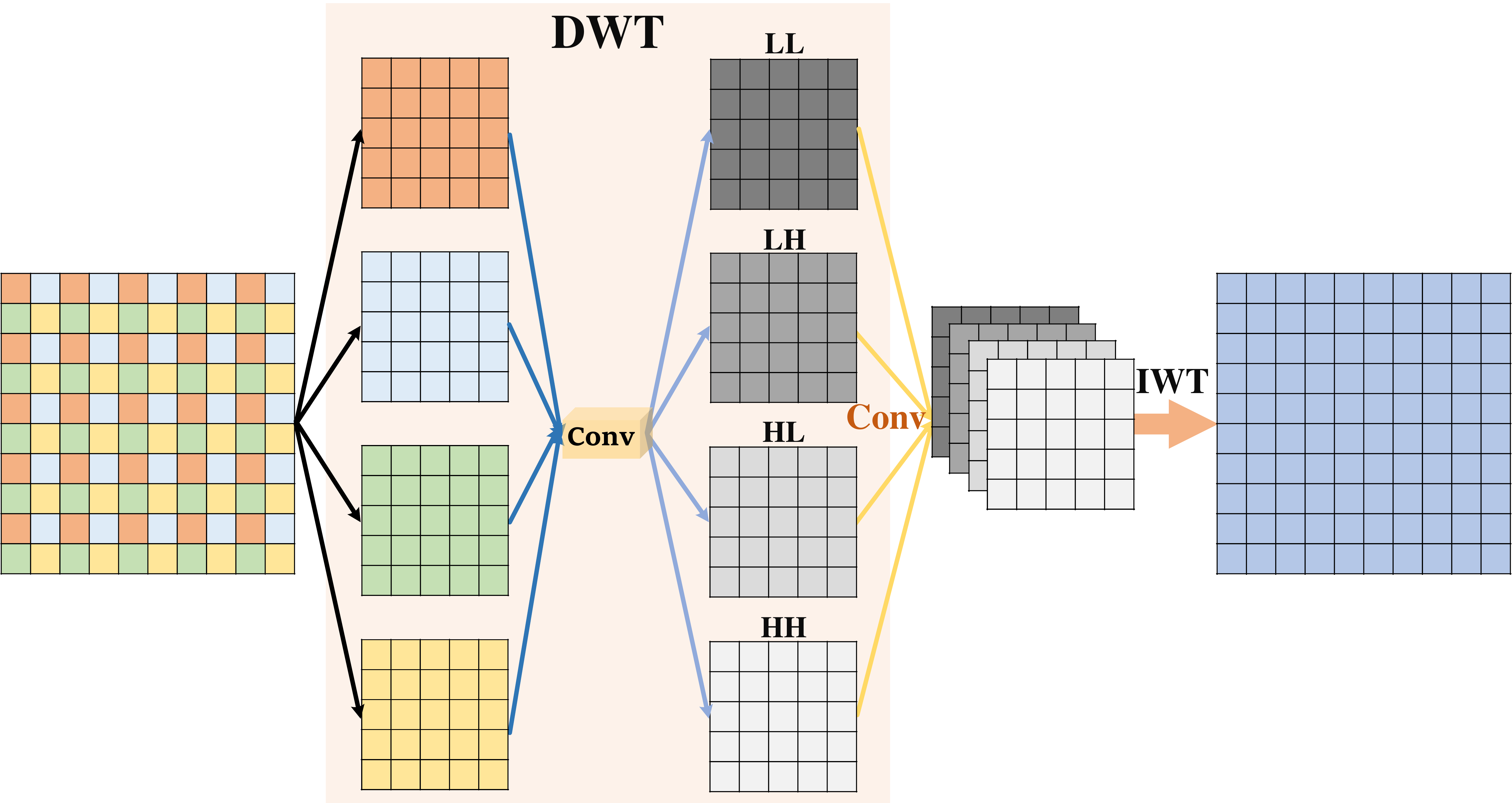}
  \label{fig:dwt}
\end{minipage}%
}
\vspace{-0ex}
\caption{\small
Illustration of average pooling, dilated filter and the proposed MWCNN. Take one CNN block as an example:
(a) sum-pooling with factor 2 leads to the most significant information loss which is not suitable for image restoration;
(b) dilated filtering with rate 2 is equal to shared parameter convolution on sub-images;
(c) the proposed MWCNN first decomposes an image into 4 sub-bands and then concatenates them as input of CNN blocks. IWT is then used as an upsampling layer to restore resolution of the image.
}
\label{fig:discussion}
\vspace{-2ex}
\end{center}}
\end{figure*}

\subsection{Network architecture}
\label{sec:arch}
\subsubsection{Image Restoration}
As mentioned previously in Sec.~\ref{sec:wpt2mw}, we design the MWCNN architecture for image restoration  based on the principle of the WPT as illustrated in Figure~\ref{fig:mwcnn_arch_g}.
The key idea is to insert CNN blocks into WPT before (or after) each level of DWT.
As shown in Figure~\ref{fig:Architecture}, each CNN block is a 3-layer FCN without pooling, and takes both low-frequency subbands and high-frequency subbands as inputs.
More concretely, each layer contains convolution with $3\! \times\! 3$ filters (Conv), and rectified linear unit (ReLU) operations. 
Only Conv is adopted in the last layer for predicting the residual result.
The number of convolutional layers is set to 24.
For more details on the setting of MWCNN, please refer to Figure \ref{fig:Architecture}.

Our MWCNN modifies U-Net in three aspects.
(i) In conventional U-Net, pooling and deconvolution are utilized as downsampling and upsampling layers.
In comparison, DWT and IWT are used in MWCNN.
(ii) After DWT, we deploy another CNN blocks to reduce the number of feature map channels for compact representation and modeling inter-band dependency.
And convolution  are adopted to increase the number of feature map channels and IWT is utilized to upsample feature map.
In comparison, conventional U-Net adopting convolution layers are used to increase feature map channels which has no effect on the number of feature map channels after pooling.
For upsampling, deconvolution layers are directly adopted to zoom in on feature map.
(iii) In MWCNN, element-wise summation is used to combine the feature maps from the contracting and expanding subnetworks.
While in conventional U-Net, concatenation is adopted.
Compared to our previous work~\cite{liu2018multi}, we have made several improvements such as:
(i)  Instead of directly decomposing input images by DWT, we first use conv blocks to extract features from input, which is empirically shown to be beneficial for image restoration.
(ii) In the 3rd hierarchical level, we use more feature maps to enhance feature representation.
In our implementation, Haar wavelet is adopted as the default wavelet in MWCNN.
Other wavelets, \textit{e.g.}, Daubechies 2 (DB2), are also considered in our experiments.

Denote by $\Theta$ the network parameters of MWCNN, \textit{i.e.}, and $\mathcal{F}\left(\mathbf{y}; \Theta \right)$ be the network output.
Let $\{\left( \mathbf{y}_i, \mathbf{x}_i \right)\}_{i=1}^N$ be a training set, where $ \mathbf{y}_i$ is the $i$-th input image, $ \mathbf{x}_i $ is the corresponding ground-truth image.
Then the objective function for learning MWCNN is  given by
\begin{equation}\label{eq:loss}
  \mathcal{L}(\Theta) = \frac{1}{2N}\sum_{i=1}^N \| \mathcal{F}(\mathbf{y}_i; \Theta) - ( \mathbf{y}_i - \mathbf{x}_i) \|_F^2.
\end{equation}
The ADAM algorithm~\cite{kingma2014adam} is adopted to train MWCNN by minimizing the objective function.

\subsubsection{Extend to Object Classification}
\label{sec:classarch}

Similar to image restoration, DWT is employed as a  downsampling operation  often without upsampling operation to replace pooling operation.
The compression filter with $1\times1$ Conv is subsequently utilized after DWT transformation.
Note that we don't modify other blocks or loss function.
With this improvement, feature can be further selected and enhanced with adaptive learning.
Moreover, any CNN using pooling can be considered instead of DWT operation, and the information of feature maps can be transmitted to next layer without information loss. 
DWT can be seen as a safe downsampling module and plugged into any CNNs without the need to  change network architectures, and may benefit extracting more powerful features for different tasks.

\subsection{Discussion}
\subsubsection{Connection to Pooling Operation}

The DWT in the proposed MWCNN is closely related to the pooling operation and dilated filtering.
By using the Haar wavelet as an example, we explain the connection between DWT and average pooling.
According to the theory of average pooling with factor 2, the $(i, j)$-th value of feature map $\mathbf{x}_{l}$ in the $l$-th layer after pooling can be written as
\begin{spacing}{0.4}
\end{spacing}
\begin{footnotesize}
\begin{equation}\label{eq:sumpooling}
\mathbf{x}_l(i,j) \!\!=\!\!  (\mathbf{x}_{l\!\!-\!\!1}(2i\!\!-\!\!1,2j\!\!-\!\!1) + \mathbf{x}_{l\!\!-\!\!1}(2i\!\!-\!\!1,2j)  + \mathbf{x}_{l\!\!-\!\!1}(2i,2j\!\!-\!\!1) + \mathbf{x}_{l\!\!-\!\!1}(2i,2j))/4,
\end{equation}
\end{footnotesize}
\noindent
where $\mathbf{x}_{l-1}$ is the feature map before pooling operation.
It is  clear that Eq.~\ref{eq:sumpooling} is the same as the low-frequency component of DWT in Eq.~\ref{eq:Haar_trans}, which also means that all the high-frequency information is lost during the pooling operation.
In Figure~\ref{fig:con2pooling}, the feature map is first decomposed into four sub-images with stride 2. 
The average pooling operation can be treated as summing all sub-images with fixed coefficient $1/4$ to  generate new sub-image.
In comparison, DWT uses all sub-images with four fixed orthometric weights to obtain four new sub-images.
By taking all the subbands into account, MWCNNs can therefore avoid the information loss caused by conventional subsampling, and may benefit restoration and classification.
Hence, average pooling can be seen as a simplified variant of the proposed MWCNNs.

\subsubsection{Connection to Dilated Filtering}

To illustrate the connection between MWCNN and dilated filtering, we first give the definition of dilated filtering with factor 2:
\begin{equation}\label{eq:dilated}
 (\mathbf{x}_l \otimes_2 \mathbf{k})(i,j) \! = \!\!\! \sum_{
  \begin{scriptsize}
  \begin{aligned}
    p\!+\!2s \!= \!i\!,\\
    q\!+\!2t \!=\! j\!
  \end{aligned}
  \end{scriptsize}
} \mathbf{x}_{l}(p,q) \mathbf{k}(s,t),
\end{equation}
where $\otimes_2$ means convolution operation with dilated factor 2, $(s,t)$ is the position in convolutional kernel $\mathbf{k}$, and $(p,q)$ is the position within the range of convolution of feature $\mathbf{x}_{l}$.
Eq.~(\ref{eq:dilated}) can be decomposed into two steps, sampling and convoluting.
Sampled patch is obtained by sampling at center position $(i, j)$ of $\mathbf{x}$ with one interval pixel under the constraint $p+2s= i, q+2t= j$. 
Then the value $\mathbf{x}_{l+1}(i,j)$ is obtained by convolving sampled patch with kernel $\mathbf{k}$.
Therefore, dilated filtering with factor 2 can be expressed as first decomposing an image into four sub-images and then using the shared standard convolutional kernel on those sub-images as illustrated  in Figure.~\ref{fig:dailated}.
We rewrite Eq.~(\ref{eq:dilated}) for obtaining the pixel value $\mathbf{x}_{l+1}(2i-1,2j-1)$ as following:
\begin{equation}\label{eq:dilated_sub}
 (\mathbf{x}_l \otimes_2 \mathbf{k})(2i-1,2j-1) \! = \!\!\! \sum_{
  \begin{scriptsize}
  \begin{aligned}
    p\!+\!2s \!= \!2i\!-\!1,\\
    q\!+\!2t \!=\! 2j\!-\!1
  \end{aligned}
  \end{scriptsize}
} \mathbf{x}_l(p,q) \mathbf{k}(s,t).
\end{equation}
Then the pixel value $\mathbf{x}_{l+1}(2i-1,2j)$, $\mathbf{x}_{l+1}(2i,2j-1)$ and $\mathbf{x}_{l+1}(2i,2j)$  can be obtained in the same way.
Actually, the value of $\mathbf{x}_l$  at the position $(2i-1,2j-1)$, $(2i-1,2j)$, $(2i,2j-1)$ and $(2i,2j)$ can be obtained by applying IWT on subband images $\mathbf{x}_l^{LL}$, $\mathbf{x}_l^{LH}$, $\mathbf{x}_l^{HL}$ and $\mathbf{x}_l^{HH}$ based on Eqn.(\ref{eq:iHaar_trans}).
Therefore, the dilating filtering can be represented as convolution with the subband images as following,
\begin{equation}\label{eq:dilated_haar}
\!\!\!\!\!\!\left\{ \!\!\!
\begin{aligned}
  (\mathbf{x}_l \! \otimes_2 \! \mathbf{k})(2i\!\! - \!\!1,2j\!\! - \!\!1)  \!\! =\! \!  \left((\mathbf{x}_l^{LL} \!\! - \!\! \mathbf{x}_l^{LH} \!\! - \!\! \mathbf{x}_l^{HL} \!\! + \!\! \mathbf{x}_l^{HH}) \! \otimes \! \mathbf{k}\right)(i,j)/4,
  \\
   (\mathbf{x}_l \otimes_2 \mathbf{k})(2i\! - \!1,2j)  \! = \!  \left((\mathbf{x}_l^{LL} \!\! - \!\! \mathbf{x}_l^{LH} \!\! + \!\! \mathbf{x}_{HL} \!\! - \!\! \mathbf{x}_l^{HH}) \! \otimes \! \mathbf{k}\right)(i,j)/4,
  \\
   (\mathbf{x}_l \otimes_2 \mathbf{k})(2i\! - \!1,2j)  \! = \!  \left((\mathbf{x}_l^{LL} \!\! + \!\! \mathbf{x}_l^{LH} \!\! - \!\! \mathbf{x}_l^{HL} \!\! - \!\! \mathbf{x}_l^{HH}) \! \otimes \! \mathbf{k}\right)(i,j)/4,
  \\
  (\mathbf{x}_l \otimes_2 \mathbf{k})(2i,2j) \!  = \!  \left((\mathbf{x}_l^{LL} \! + \! \mathbf{x}_l^{LH} \! + \! \mathbf{x}_l^{HL} \! + \! \mathbf{x}_l^{HH})  \otimes \mathbf{k}\right)(i,j)/4.
\end{aligned}
\right.
\end{equation}
%

Different from dilated filtering, the definition of MWCNN in Figure~\ref{fig:dwt} can be given as \[ \mathbf{x}_l^{DWT} \otimes \mathbf{k}, \] where $\mathbf{x}_l^{DWT}=Concat(\mathbf{x}_l^{LL}, \mathbf{x}_l^{LH}, \mathbf{x}_l^{HL}, \mathbf{x}_l^{HH})$, and $Concat(\cdot,\cdot)$ denotes concatenate operation.
If $\mathbf{k}$ is group convolution~\cite{xie2017aggregated} with factor 4, the equation can be rewritten as:
\begin{equation}\label{eq:mw}
\left(\mathbf{x}_l^{DWT} \otimes \mathbf{k}\right) (i, j)  = \!\!\!\!\!\! \sum_{i=\{LL, LH, HL, HH \} } \!\!\!\!\!\!\!\!\!\!\!\! \left( \mathbf{x}_l^{i} \otimes \mathbf{k}_i\right) \left(i, j\right).
\end{equation}
Note that $\left( \mathbf{x}_l^{LL}, \mathbf{x}_l^{LH}, \mathbf{x}_l^{HL}, \mathbf{x}_l^{HH} \right)$ can accurately reconstruct $\mathbf{x}_l$ by using IWT.
Compared to Eq.~(\ref{eq:dilated_haar}), the weights of each subband $\mathbf{x}_l^{i}$ and the corresponding convolution $\mathbf{k}_i$ are different.
That means that our MWCNN can be reduced to dilated filtering if the subbands $\left( \mathbf{x}_l^{LL}, \mathbf{x}_l^{LH}, \mathbf{x}_l^{HL}, \mathbf{x}_l^{HH} \right)$ are replaced by subimages after IWT in Eq.~(\ref{eq:iHaar_trans}), and the convolution $\mathbf{k}_i$ in $\mathbf{k}$ is shared to each other.
Hence, the dilated filtering can be seen as a variant of the proposed MWCNN. 

\begin{figure}[!htbp]
\scriptsize{

\begin{center}
\subfigure[]{
\begin{minipage}[c]{0.155\textwidth}
\centering
  \includegraphics[width=0.99\linewidth]{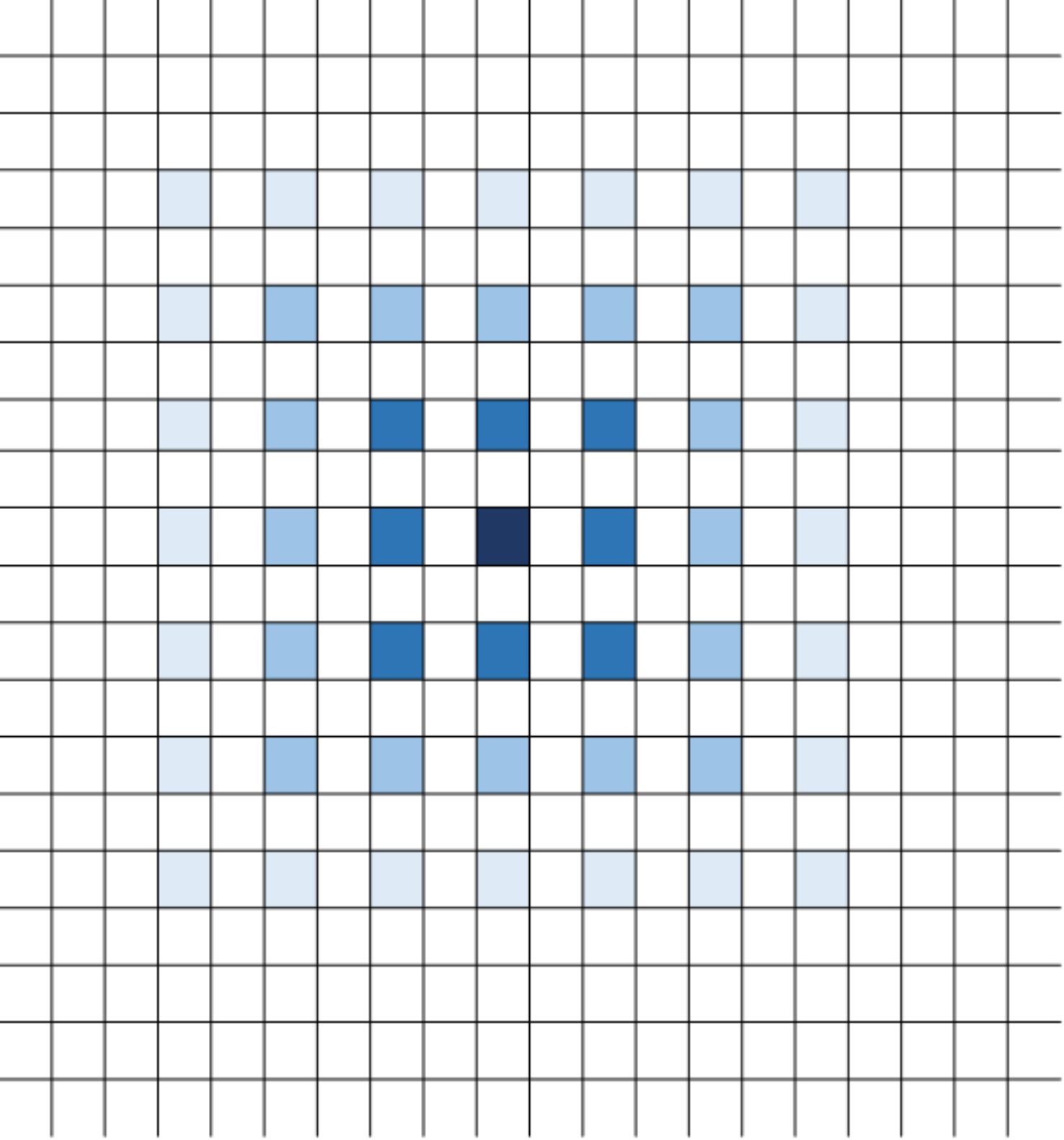}
  \label{fig:RF_d}
\end{minipage}%
}
\hspace{-1ex}
\subfigure[]{
\begin{minipage}[c]{0.155\textwidth}
\centering
  \includegraphics[width=0.99\linewidth]{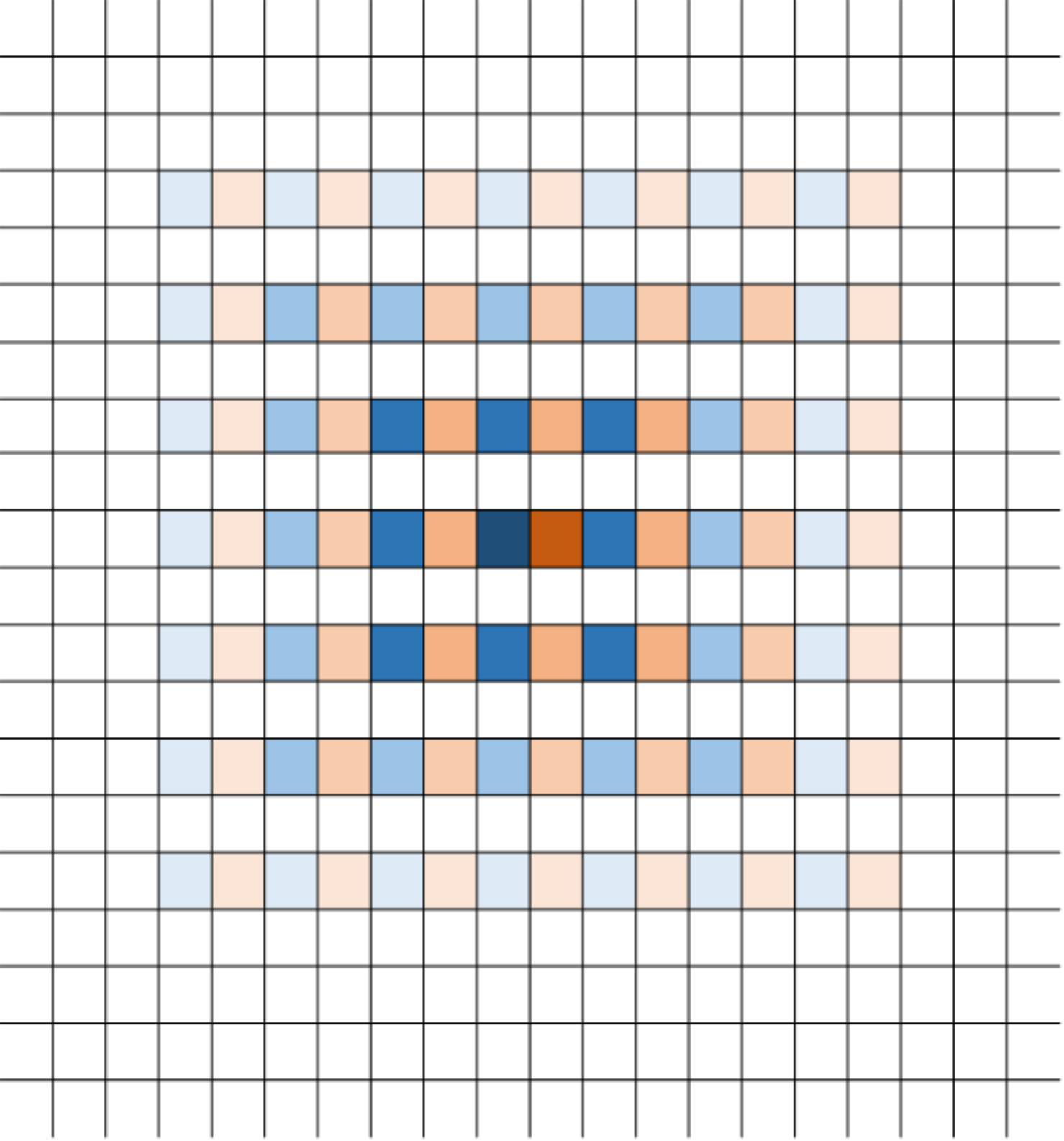}
  \label{fig:RF2D}
\end{minipage}%
}
\hspace{-1ex}
\subfigure[]{
\begin{minipage}[c]{0.155\textwidth}
\centering
  \includegraphics[width=0.99\linewidth]{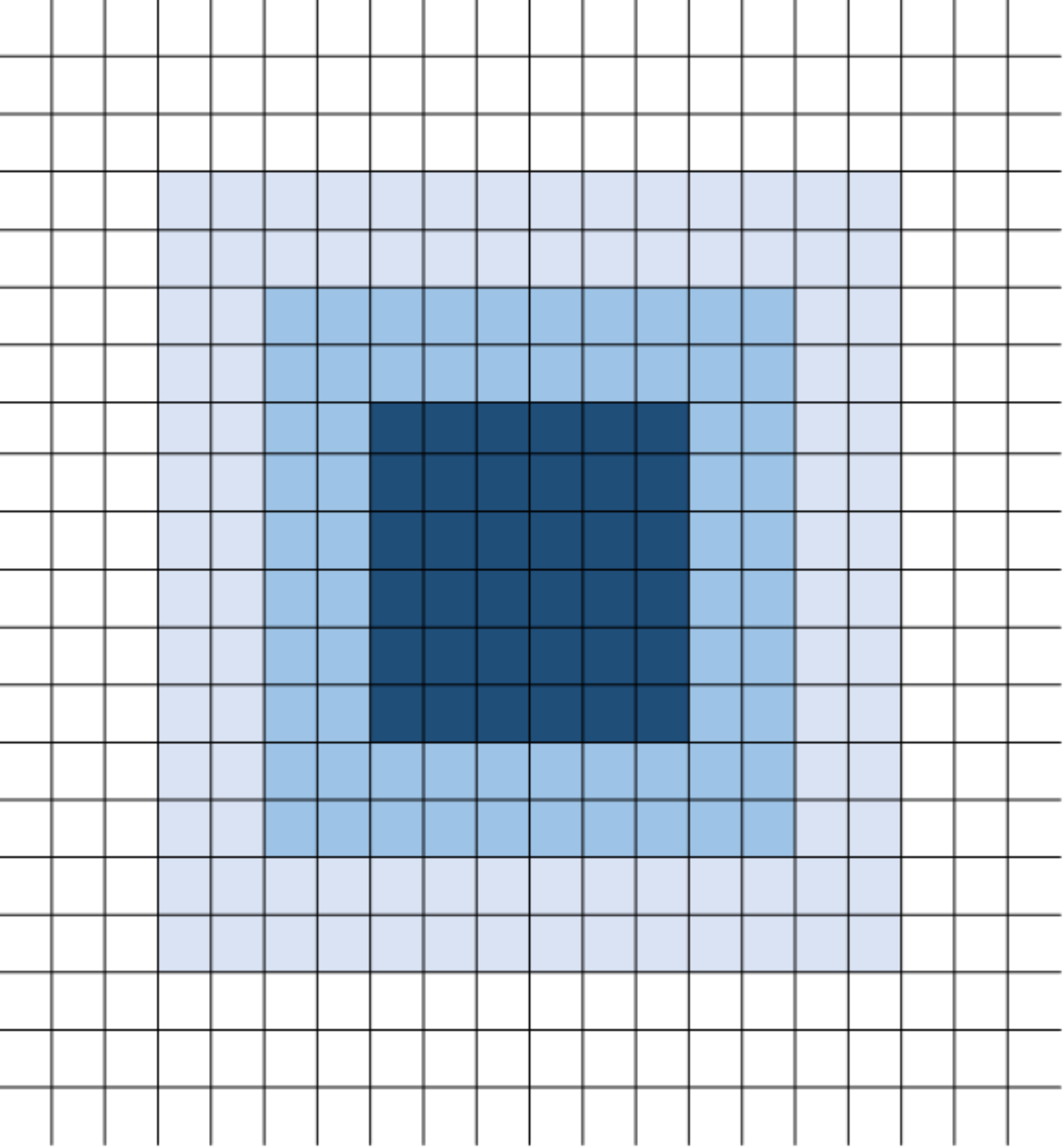}
  \label{fig:RFMW}
\end{minipage}%
}
\vspace{-2ex}
\caption{\small Illustration of the gridding effect.
         Taken 3-layer CNNs as an example: (a) the dilated filtering with rate 2 suffers from large amount of information loss,  (b)  the two neighbored pixels are based on information from totally non-overlapped locations, and (c) our MWCNN can perfectly avoid underlying drawbacks.}\label{fig:gridding}
\vspace{-0ex}
\end{center}}
\end{figure}

Compared with dilated filtering, MWCNN can also avoid the gridding effect.
With the increase of depth, dilated filtering with fixed factor greater than 1 only considers a sparse sampling of units in the checkerboard pattern, resulting in large amount of information loss (see Figure~\ref{fig:RF_d}).
Another problem with dilated filtering is that the two output neighboring pixels may be computed from input information from totally non-overlapped units (see Figure~\ref{fig:RF2D}), and may cause the inconsistence of local information.
Figure~\ref{fig:RFMW} illustrates the receptive field of MWCNN, which is quite different from dilated filtering.
With dense sampling, convolution filter takes multi-frequency information as input, and results in double receptive field after DWT.
One can see that MWCNN is able to well address the sparse sampling and inconsistency problems of local information, and is expected to benefit restoration
quantitatively. 


\section{Experiments}\label{sec:exp}
\label{sec:exp}

In this section, we first describe application of MWCNN to image restoration.
Then ablation experiments  is presented to analyze the contribution of each component.
Finally, the proposed MWCNN is extended to object classification.


\subsection{Experimental Setting for Image Restoration}

\begin{table*}[!htbp]\footnotesize
\vspace{-0.01in}
\centering
\caption{Average PSNR(dB)/SSIM results of the competing methods for image denoising with noise levels $\sigma = $ 15, 25 and 50 on datasets Set14, BSD68 and Urban100. Red color indicates the best performance.}
\renewcommand\tabcolsep{1.5pt}
\renewcommand\arraystretch{0.99}
{
\begin{tabular}{ p{1.1cm}<{\centering} p{0.4cm}<{\centering} p{1.72cm}<{\centering} p{1.72cm}<{\centering} p{1.72cm}<{\centering} p{1.72cm}<{\centering} p{1.72cm}<{\centering} p{1.72cm}<{\centering} p{1.72cm}<{\centering} p{1.72cm}<{\centering} p{1.72cm}<{\centering} }
\hlineB{2.2}
Dataset &   $\sigma$ & BM3D~\cite{dabov2007image} & TNRD~\cite{Chen2015Trainable} & DnCNN~\cite{Zhang2016Beyond} & IRCNN~\cite{zhang2017learning} & RED30~\cite{Mao2016Image}  & MemNet~\cite{tai2017memnet} &    FFDNet~\cite{zhang2018ffdnet} &   MWCNN(P)   &    MWCNN    \\
\hlineB{1.2}
 \multirow{3}{*}{Set12}
&15 & 32.37 / 0.8952 & 32.50 / 0.8962 & 32.86 / 0.9027 & 32.77 / 0.9008 &  -  &  - &  32.75 / 0.9027 & 33.15 / 0.9088 & \textcolor[rgb]{1,0,0}{33.20} / \textcolor[rgb]{1,0,0}{0.9089} \\
&25 & 29.97 / 0.8505 & 30.05 / 0.8515 & 30.44 / 0.8618 & 30.38 / 0.8601 &  -  &  - &  30.43 / 0.8634 & 30.79 / 0.8711 & \textcolor[rgb]{1,0,0}{30.84} / \textcolor[rgb]{1,0,0}{0.8718} \\
&50 & 26.72 / 0.7676 & 26.82 / 0.7677 & 27.18 / 0.7827 & 27.14 / 0.7804 & 27.34 / 0.7897 & 27.38 / 0.7931 & 27.32 / 0.7903  & 27.74 / 0.8056 & \textcolor[rgb]{1,0,0}{27.79} / \textcolor[rgb]{1,0,0}{0.8060} \\ \hline
 \multirow{3}{*}{BSD68}
&15 & 31.08 / 0.8722 & 31.42 / 0.8822 & 31.73 / 0.8906 & 31.63 / 0.8881 &  -  & -  & 31.63 / 0.8902 & 31.86 / 0.8947 & \textcolor[rgb]{1,0,0}{31.91} / \textcolor[rgb]{1,0,0}{0.8952} \\
&25 & 28.57 / 0.8017 & 28.92 / 0.8148 & 29.23 / 0.8278 & 29.15 / 0.8249 &  -  & -  & 29.19 / 0.8289 & 29.41 / 0.8360 & \textcolor[rgb]{1,0,0}{29.46} / \textcolor[rgb]{1,0,0}{0.8370} \\
&50 & 25.62 / 0.6869 & 25.97 / 0.7021 & 26.23 / 0.7189 & 26.19 / 0.7171 &  26.35 / 0.7245  &  26.35 / 0.7294 & 26.29 / 0.7245  & 26.53 / 0.7366 & \textcolor[rgb]{1,0,0}{26.58} / \textcolor[rgb]{1,0,0}{0.7382}  \\ \hline
 \multirow{3}{*}{Urban100}
&15 & 32.34 / 0.9220 & 31.98 / 0.9187 & 32.67 / 0.9250 & 32.49 / 0.9244 & -  & - & 32.43 / 0.9273 & 33.17 / 0.9357 & \textcolor[rgb]{1,0,0}{33.22} / \textcolor[rgb]{1,0,0}{0.9361} \\
&25 & 29.70 / 0.8777 & 29.29 / 0.8731 & 29.97 / 0.8792 & 29.82 / 0.8839 & -  & - & 29.92 / 0.8886  & 30.66 / 0.9026 & \textcolor[rgb]{1,0,0}{30.74} / \textcolor[rgb]{1,0,0}{0.9035} \\
&50 & 25.94 / 0.7791 & 25.71 / 0.7756 & 26.28 / 0.7869 & 26.14 / 0.7927 &  26.48 / 0.7991  & 26.64 / 0.8024 & 26.52 / 0.8057  & 27.42 / 0.8371 & \textcolor[rgb]{1,0,0}{27.53} / \textcolor[rgb]{1,0,0}{0.8393} \\
\hlineB{2.2}
\end{tabular}
}
\vspace{-0.11in}
\label{tab:denoising}
\end{table*}
\begin{figure*}[!htbp]
\setlength{\abovecaptionskip}{0pt}
\setlength{\belowcaptionskip}{0pt}
  \centering

\subfigure{
\begin{minipage}[c]{0.22\textwidth}
\centering
  \includegraphics[width=1\textwidth,height=0.72\textwidth]{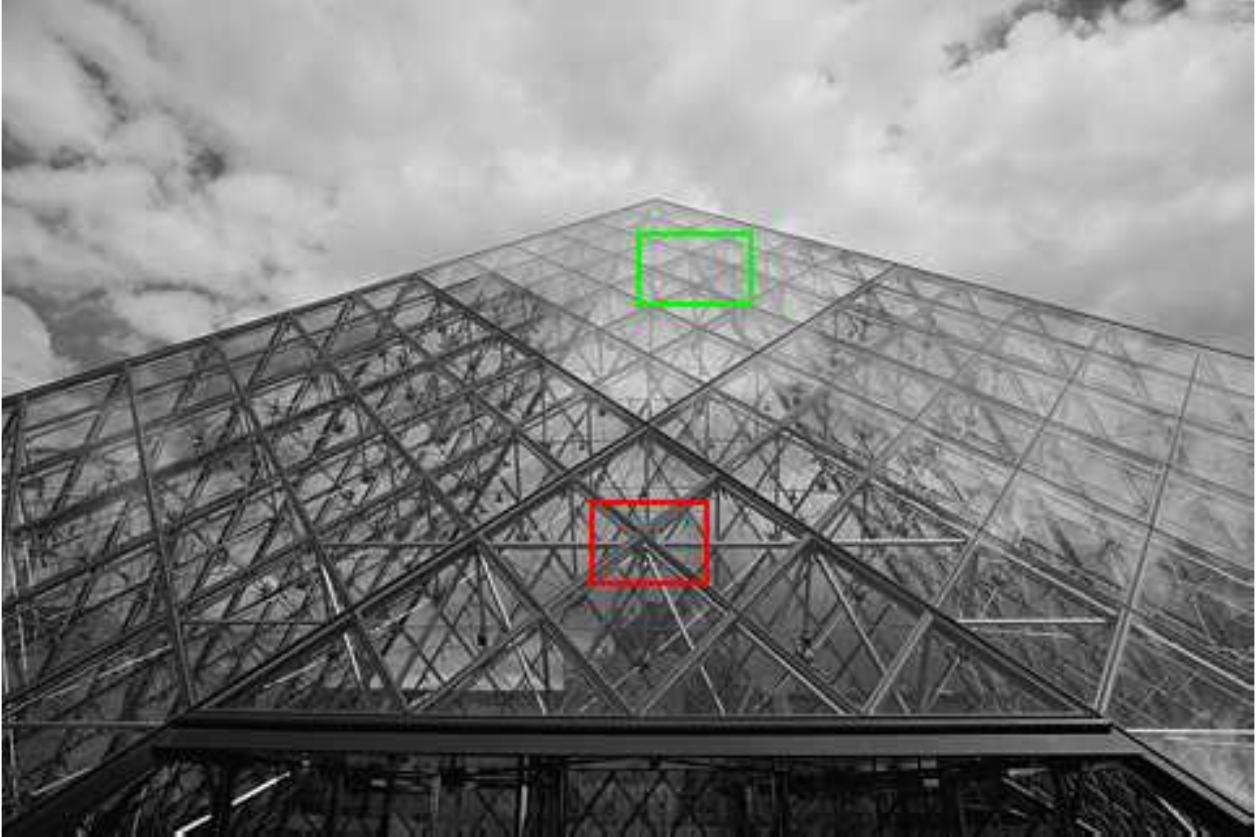}
  \centerline{ Ground Truth}
\end{minipage}%
}
\begin{minipage}[c]{0.765\textwidth}
\vspace{-0.00in}
\subfigure{
\begin{minipage}[c]{0.195\textwidth}
\centering
  \includegraphics[width=.49\textwidth]{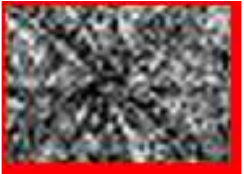}\hspace{-0.04in}
  \includegraphics[width=.49\textwidth]{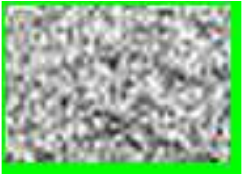}
  \centerline{\scalebox{0.99}{ Noisy Image  }}
\end{minipage}%
}
\hspace{-0.1in}
\subfigure{
\begin{minipage}[c]{0.195\textwidth}
\centering
  \includegraphics[width=.49\textwidth]{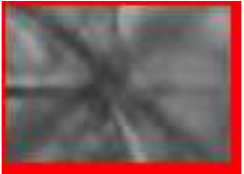}\hspace{-0.04in}
  \includegraphics[width=.49\textwidth]{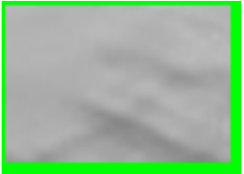}
  \centerline{\scalebox{0.99}{ BM3D~\cite{dabov2007image}  }}
\end{minipage}%
}
\hspace{-0.1in}
\subfigure{
\begin{minipage}[c]{0.195\textwidth}
\centering
  \includegraphics[width=.49\textwidth]{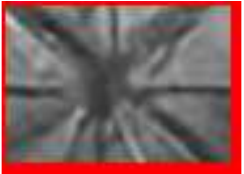}\hspace{-0.04in}
  \includegraphics[width=.49\textwidth]{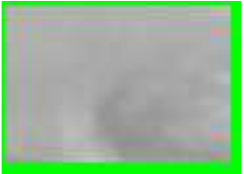}
  \centerline{\scalebox{0.99}{ TNRD~\cite{Chen2015Trainable}  }}
\end{minipage}%
}
\hspace{-0.1in}
\subfigure{
\begin{minipage}[c]{0.195\textwidth}
\centering
  \includegraphics[width=.49\textwidth]{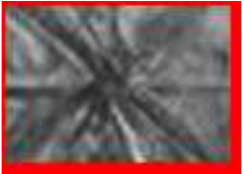}\hspace{-0.04in}
  \includegraphics[width=.49\textwidth]{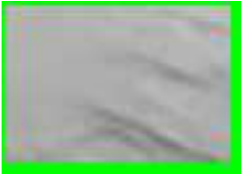}
  \centerline{\scalebox{0.99}{ DnCNN~\cite{Zhang2016Beyond}  }}
\end{minipage}%
}
\hspace{-0.1in}
\subfigure{
\begin{minipage}[c]{0.195\textwidth}
\centering
  \includegraphics[width=.49\textwidth]{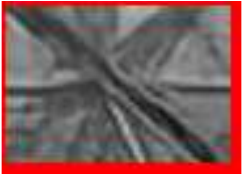}\hspace{-0.04in}
  \includegraphics[width=.49\textwidth]{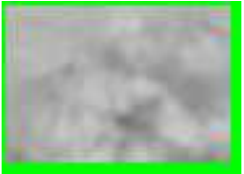}
  \centerline{\scalebox{0.99}{ IRCNN~\cite{zhang2017learning}  }}
\end{minipage}%
}

\vspace{-0.01in}
\subfigure{
\begin{minipage}[c]{0.195\textwidth}
\centering
  \includegraphics[width=.49\textwidth]{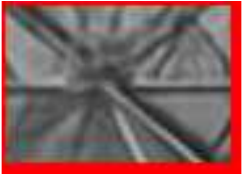}\hspace{-0.04in}
  \includegraphics[width=.49\textwidth]{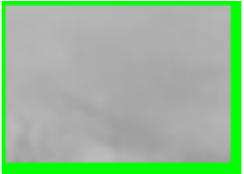}
  \centerline{\scalebox{0.99}{ RED30~\cite{Mao2016Image}  }}
\end{minipage}%
}
\hspace{-0.1in}
\subfigure{
\begin{minipage}[c]{0.195\textwidth}
\centering
  \includegraphics[width=.49\textwidth]{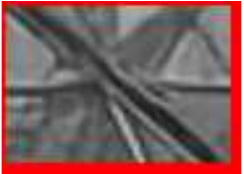}\hspace{-0.04in}
  \includegraphics[width=.49\textwidth]{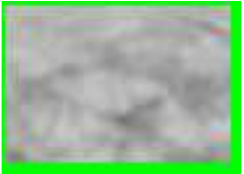}
  \centerline{\scalebox{0.99}{ MemNet~\cite{tai2017memnet}  }}
\end{minipage}%
}
\hspace{-0.1in}
\subfigure{
\begin{minipage}[c]{0.195\textwidth}
\centering
  \includegraphics[width=.49\textwidth]{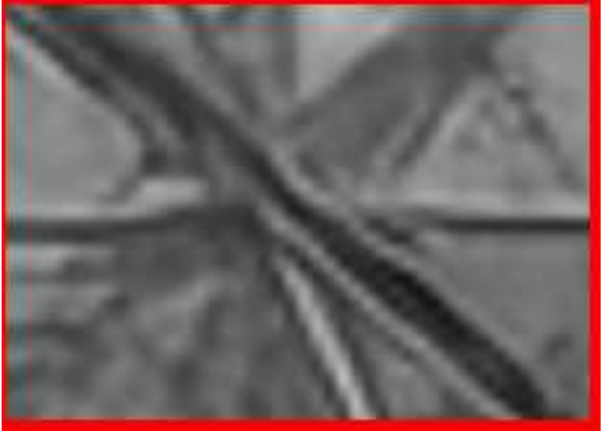}\hspace{-0.04in}
  \includegraphics[width=.49\textwidth]{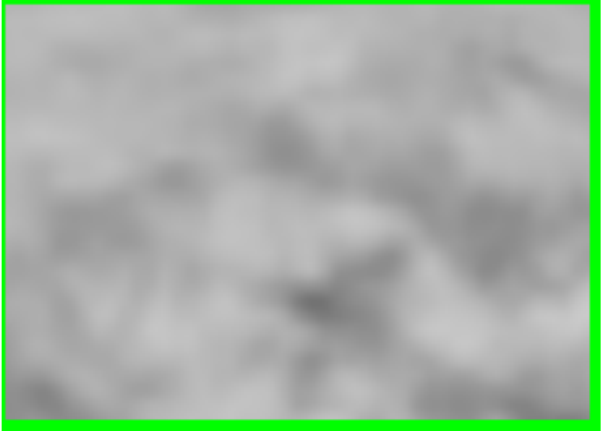}
  \centerline{\scalebox{0.99}{ FFDNet~\cite{zhang2018ffdnet}  }}
\end{minipage}%
}
\hspace{-0.1in}
\subfigure{
\begin{minipage}[c]{0.195\textwidth}
\centering
  \includegraphics[width=.49\textwidth]{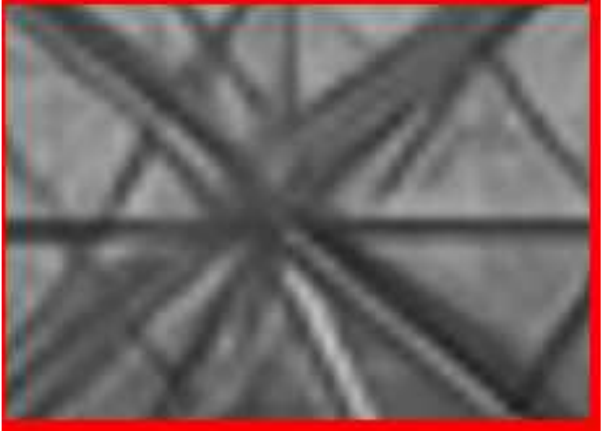}\hspace{-0.04in}
  \includegraphics[width=.49\textwidth]{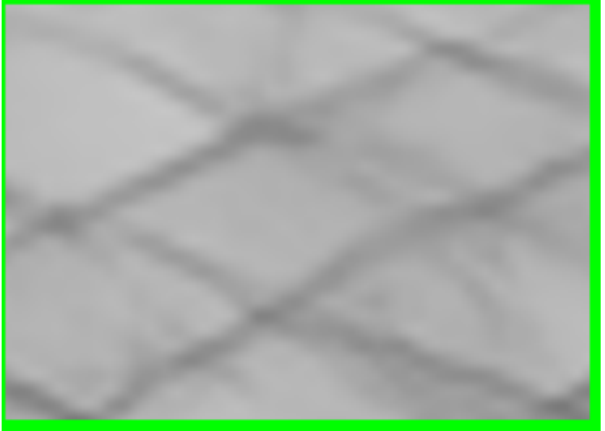}
  \centerline{\scalebox{0.99}{ MWCNN }}
\end{minipage}%
}
\hspace{-0.1in}
\subfigure{
\begin{minipage}[c]{0.195\textwidth}
\centering
  \includegraphics[width=.49\textwidth]{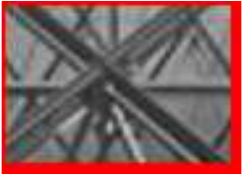}\hspace{-0.04in}
  \includegraphics[width=.49\textwidth]{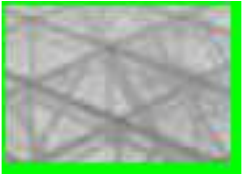}
  \centerline{\scalebox{0.99}{ Ground Truth  }}
\end{minipage}%
}

\vspace{-0.01in}
\end{minipage}
\caption{\textit{Image denoising} results of ``$Test044$'' (Set68) with noise level of 50. }\label{fig:DN50_test044}
\vspace{-0.01in}
\end{figure*}

\subsubsection{Training set}

To train our MWCNN, we adopt DIV2K~\cite{agustsson2017ntire} as our training dataset.
Concretely, DIV2K contains $800$ images with about 2K resolution for training, $100$ images for validation, and $100$ images for testing.
Due to the receptive field of MWCNN being $181 \times 181$, we crop $N = 20 \times 4,000$  patches with the size of $192\times 192$ from the training images in the training stage.

For image denoising, we consider three noise levels, \textit{i.e.}, $\sigma$ = 15, 25 and 50, and evaluate our denoising method on three dataset, \textit{i.e.}, Set12~\cite{Zhang2016Beyond}, BSD68~\cite{MartinFTM01}, and Urban100~\cite{huang2015single}.
For SISR, we take $bicubic$ upsampling as the input to MWCNN with three specific scale factors, \textit{i.e.}, $\times2$, $\times3$ and $\times4$, respectively. 
Four widely used datasets, Set5~\cite{bevilacqua2012low}, Set14~\cite{zeyde2010single}, BSD100~\cite{MartinFTM01} and Urban100~\cite{huang2015single}, are adopted to evaluate SISR performance.
For JPEG image artifacts removal, we follow the setting as used in \cite{Dong2016Compression}, \textit{i.e.}, four compression quality settings $Q$ = 10, 20, 30 and 40 for the JPEG encoder.
Two datasets, Classic5~\cite{Dong2016Compression} and LIVE1~\cite{moorthy2009visual}, are used for evaluating our method.

\subsubsection{Network training}

In image restoration, a MWCNN model is learned for each degradation setting.
The ADAM algorithm \cite{kingma2014adam} with $\beta_1=0.9$, $\beta_2=0.999$ and $\epsilon=10^{-8}$ is adopted for optimization and we use a mini-batch size of 24.
The learning rate is decayed exponentially from $10^{-4}$ to $10^{-5}$ in the 200 epochs.
Rotation or/and flip based data augmentation is used during mini-batch learning.
The MatConvNet package \cite{vedaldi2015matconvnet} with NVIDIA GTX1080 GPU is utilized for training and testing.

\subsection{Quantitative and qualitative evaluation on Image Restoration Tasks}

Comprehensive experiments are conducted to evaluate our 24-layer MWCNN using the same setting as in Sec.~\ref{sec:arch} on three representative image restoration tasks, respectively.
Here, we also provide the results of our previous work and denote it as MWCNN(P)~\cite{liu2018multi}.

\subsubsection{Image denoising}

For image denoising, only gray images are trained and evaluated for the reason that most denoising methods are only trained and tested on gray images.
Moreover, we compare with two classic denoising methods, \textit{i.e.}, BM3D~\cite{dabov2007image} and TNRD~\cite{Chen2015Trainable}, and five CNN-based methods, \textit{i.e.}, DnCNN~\cite{Zhang2016Beyond}, IRCNN~\cite{zhang2017learning}, RED30~\cite{Mao2016Image}, MemNet~\cite{tai2017memnet}, and FFDNet~\cite{zhang2018ffdnet}.
Table \ref{tab:denoising} lists the average PSNR/SSIM results of the competing methods on these three datasets.
Since RED30~\cite{Mao2016Image}and MemNet~\cite{tai2017memnet} doesn't train the models on level 15 and level 25, we use the symbol `-' instead.
Obviously, the performance of all the competing methods are worse than our MWCNN.
It's worth noting that our MWCNN can outperform DnCNN and FFDNet by about $0.3\sim0.5$dB in terms of PSNR on Set12, and slightly surpass with $0.2\sim0.3$dB on BSD68.
On Urban100, our MWCNN generally achieves favorable performance when compared with the competing methods.
Specially, the average PSNR by our MWCNN can be 0.5dB higher than that by DnCNN on Set12, and 1.2dB higher on Urban100 when the noise level $\sigma$ is 50.
Figure~\ref{fig:DN50_test044} shows the denoising results of the images  ``\textit{Test044}'' from Set68 with the noise level $\sigma = 50$.
One can see that our MWCNN is promising in removing noise while recovering image details and structures, and can obtain visually more pleasant result than the competing methods due to the reversibility of WPT during downsampling and upsampling.

\begin{table*}[!htbp]\footnotesize
\centering
\vspace{-0.01in}
\caption{Average PSNR(dB) / SSIM results of  the competing methods for SISR with scale factors $S=$ 2, 3 and 4 on datasets Set5, Set14, BSD100 and Urban100. Red color indicates the best performance.}
\renewcommand\tabcolsep{1.7pt}
\renewcommand\arraystretch{0.99}
\scalebox{0.800}{
\begin{tabular}{p{1.2cm}<{\centering} p{0.7cm}<{\centering} p{1.72cm}<{\centering} p{1.72cm}<{\centering} p{1.72cm}<{\centering} p{1.72cm}<{\centering} p{1.72cm}<{\centering} p{1.72cm}<{\centering} p{1.72cm}<{\centering} p{2.02cm}<{\centering} p{1.72cm}<{\centering} p{1.72cm}<{\centering} p{1.72cm}<{\centering} p{1.62cm}<{\centering}}
\hlineB{2.2}
Dataset& $S$ & VDSR~\cite{Kim2015Accurate}  & DnCNN~\cite{Zhang2016Beyond} & RED30~\cite{Mao2016Image} & SRResNet~\cite{Ledig2017Photo} & LapSRN~\cite{lai2017deep} & DRRN~\cite{Tai2017Image}   & MemNet~\cite{tai2017memnet} & WaveResNet~\cite{bae2017beyond} & SRMDNF~\cite{Zhang_2018_CVPR} &   MWCNN(P)  &   MWCNN \\
\hlineB{1.2}
 \multirow{3}{*}{Set5}
 &$\times$2       & 37.53 / 0.9587     &   37.58 / 0.9593  & 37.66 / 0.9599  &      -         & 37.52 / 0.9590 & 37.74 / 0.9591 & 37.78 / 0.9597   & 37.57 / 0.9586 &  37.79 / 0.9601 & 37.91 / 0.9600 & \textcolor[rgb]{1,0,0}{37.95} / \textcolor[rgb]{1,0,0}{0.9605} \\
 &$\times$3       &   33.66 / 0.9213   &   33.75 / 0.9222  & 33.82 / 0.9230  &      -         &  -  & 34.03 / 0.9244 & 34.09 / 0.9248  & 33.86 / 0.9228  & 34.12 / 0.9250 & 34.18 / 0.9272 & \textcolor[rgb]{1,0,0}{34.21} /  \textcolor[rgb]{1,0,0}{0.9273}  \\
 &$\times$4      &   31.35 / 0.8838   &   31.40 / 0.8845  & 31.51 / 0.8869 & 32.05 / 0.8902 & 31.54 / 0.8850 & 31.68 / 0.8888 &  31.74 / 0.8893 & 31.52 / 0.8864  & 31.96 / 0.8925 & 32.12 / 0.8941 & \textcolor[rgb]{1,0,0}{32.14} /  \textcolor[rgb]{1,0,0}{0.8951} \\
  \hline
   \multirow{3}{*}{Set14}
 &$\times$2       & 33.03 / 0.9124   &   33.04 / 0.9118  & 32.94 / 0.9144  &      -         & 33.08 / 0.9130 & 33.23 / 0.9136 &  33.28 / 0.9142   & 33.09 / 0.9129  & 33.05 / 0.8985 & 33.70 / 0.9182 & \textcolor[rgb]{1,0,0}{33.71} / \textcolor[rgb]{1,0,0}{0.9182} \\
 &$\times$3       & 29.77 / 0.8314   &   29.76 / 0.8349  & 29.61 / 0.8341  &      -         &       -        & 29.96 / 0.8349 & 30.00 / 0.8350  &  29.88 / 0.8331  &  30.04 / 0.8372   & 30.16 / 0.8414 & \textcolor[rgb]{1,0,0}{30.14} / \textcolor[rgb]{1,0,0}{0.8413} \\
 &$\times$4      & 28.01 / 0.7674   &   28.02 / 0.7670  & 27.86 / 0.7718   & 28.49 / 0.7783 & 28.19 / 0.7720 & 28.21 / 0.7720 &  28.26 / 0.7723  & 28.11 / 0.7699  & 28.41 / 0.7816 & 28.41 / 0.7816 & \textcolor[rgb]{1,0,0}{28.58} / \textcolor[rgb]{1,0,0}{0.7882} \\ \hline
 \multirow{3}{*}{BSD100}
 &$\times$2    & 31.90 / 0.8960   &   31.85 / 0.8942  & 31.98 / 0.8974 &      -         &  31.80 / 0.8950  & 32.05 / 0.8973 &  32.08 / 0.8978  & 31.92 / 0.8965 &  32.23 / 0.8999 & 32.23 / 0.8999 & \textcolor[rgb]{1,0,0}{32.30} / \textcolor[rgb]{1,0,0}{0.9002} \\
 &$\times$3    & 28.82 / 0.7976   &   28.80 / 0.7963  & 28.92 / 0.7993 &      -         &  - & 28.95 / 0.8004 & 28.96 / 0.8001& 28.86 / 0.7987 &  28.97 / 0.8030 & 29.12 / 0.8060 & \textcolor[rgb]{1,0,0}{29.18} / \textcolor[rgb]{1,0,0}{0.8106} \\
 &$\times$4    & 27.29 / 0.7251   &   27.23 / 0.7233  & 27.39 / 0.7286 & 27.56 / 0.7354 &  27.32 / 0.7280 & 27.38 / 0.7284 & 27.40 / 0.7281 & 27.32 / 0.7266 & 27.62 / 0.7355 & 27.62 / 0.7355 & \textcolor[rgb]{1,0,0}{27.67} / \textcolor[rgb]{1,0,0}{0.7357} \\
\hline
\multirow{3}{*}{Urban100}
 &$\times$2   &30.76 / 0.9140 &     30.75 / 0.9133 & 30.91 / 0.9159 &      -         &  30.41 / 0.9100 & 31.23 / 0.9188 & 31.31 / 0.9195 & 30.96 / 0.9169& 32.30 / 0.9296  & 32.30 / 0.9296 &  \textcolor[rgb]{1,0,0}{32.36} /  \textcolor[rgb]{1,0,0}{0.9306} \\
 &$\times$3   &27.14 / 0.8279 &     27.15 / 0.8276 & 27.31 / 0.8303 &      -         &  - &  27.53 / 0.8378 & 27.56 / 0.8376  & 27.28 / 0.8334 &  27.57 / 0.8401  & 28.13 / 0.8514 & \textcolor[rgb]{1,0,0}{28.19} /  \textcolor[rgb]{1,0,0}{0.8520} \\
 &$\times$4   &25.18 / 0.7524 &     25.20 / 0.7521 & 25.35 / 0.7587 & 26.07 / 0.7839 & 25.21 / 0.7560 & 25.44 / 0.7638 & 25.50 / 0.7630   & 25.36 / 0.7614 & 26.27 / 0.7890 & 26.27 / 0.7890 & \textcolor[rgb]{1,0,0}{26.37} / \textcolor[rgb]{1,0,0}{0.7891} \\

\hlineB{2.2}
\end{tabular}
}
\vspace{-0.11in}
\label{tab:sr}
\end{table*}
\begin{figure*}[!htbp]
\setlength{\abovecaptionskip}{0pt}
\setlength{\belowcaptionskip}{0pt}
  \centering


\subfigure{
\begin{minipage}[c]{0.25\textwidth}
\centering
  \includegraphics[width=1\textwidth,height=0.6\textwidth]{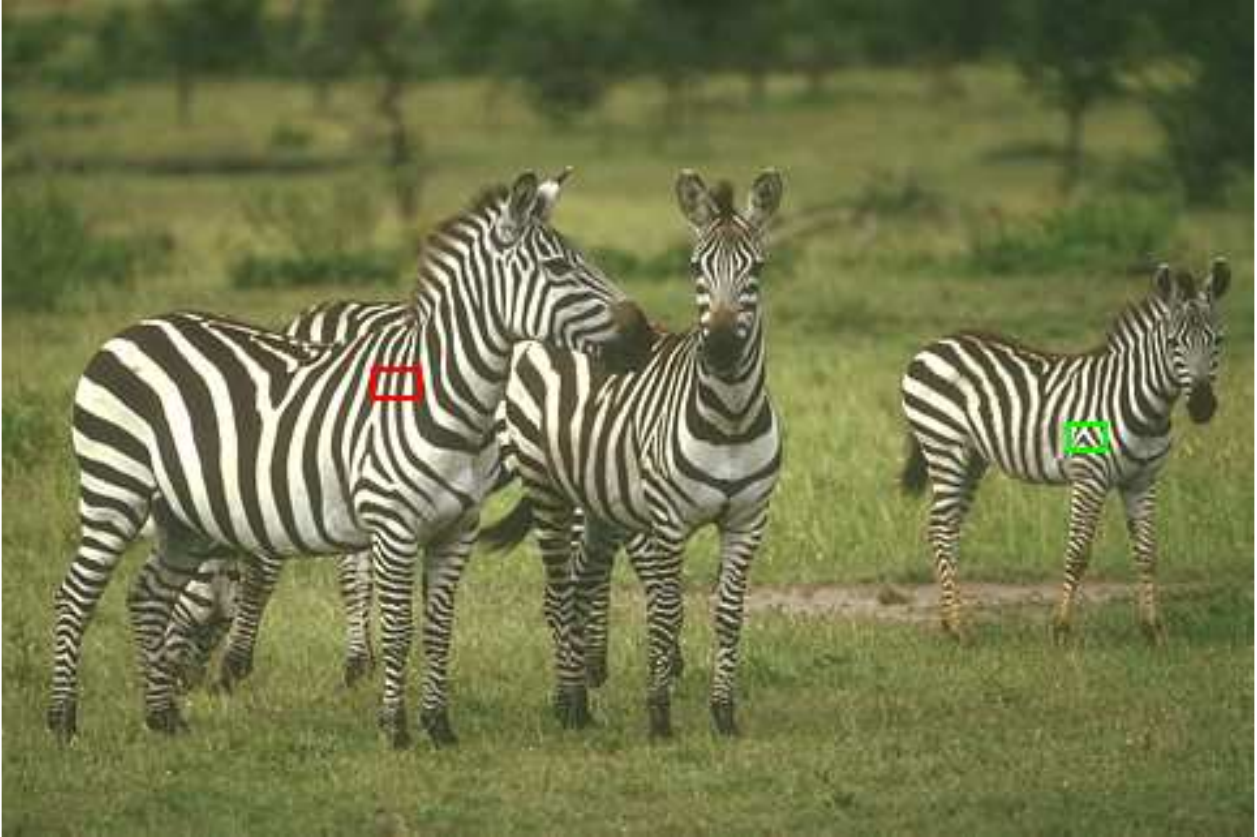}

  \centerline{ Ground Truth}
\end{minipage}%
}
\begin{minipage}[c]{0.735\textwidth}
\hspace{-0.08in}
\subfigure{
\begin{minipage}[c]{0.19\textwidth}
\centering
  \includegraphics[width=.49\textwidth]{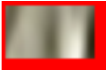}\hspace{-0.03in}
  \includegraphics[width=.49\textwidth]{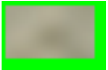}
  \centerline{\scalebox{0.99}{ VDSR~\cite{Kim2015Accurate}}}
\end{minipage}%
}
\hspace{-0.06in}
\subfigure{
\begin{minipage}[c]{0.19\textwidth}
\centering
  \includegraphics[width=.49\textwidth]{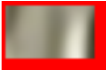}\hspace{-0.03in}
  \includegraphics[width=.49\textwidth]{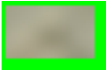}
  \centerline{\scalebox{0.99}{ DnCNN~\cite{Zhang2016Beyond}}}
\end{minipage}%
}
\hspace{-0.05in}
\subfigure{
\begin{minipage}[c]{0.19\textwidth}
\centering
  \includegraphics[width=.49\textwidth]{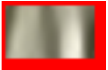}\hspace{-0.03in}
  \includegraphics[width=.49\textwidth]{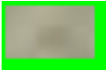}
  \centerline{\scalebox{0.99}{ RED30~\cite{Mao2016Image}}}
\end{minipage}%
}
\vspace{0.04in}
\hspace{-0.05in}
\subfigure{
\begin{minipage}[c]{0.19\textwidth}
\centering
  \includegraphics[width=.49\textwidth]{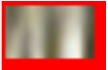}\hspace{-0.03in}
  \includegraphics[width=.49\textwidth]{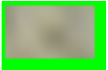}
  \centerline{\scalebox{0.99}{ SRResNet~\cite{Ledig2017Photo} }}
\end{minipage}%
}
\hspace{-0.05in}
\subfigure{
\begin{minipage}[c]{0.19\textwidth}
\centering
  \includegraphics[width=.49\textwidth]{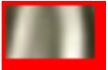}\hspace{-0.03in}
  \includegraphics[width=.49\textwidth]{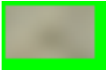}
  \centerline{\scalebox{0.99}{ LapSRN~\cite{lai2017deep} }}
\end{minipage}%
}

\hspace{-0.08in}
\subfigure{
\begin{minipage}[c]{0.19\textwidth}
\centering
  \includegraphics[width=.49\textwidth]{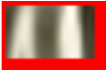}\hspace{-0.03in}
  \includegraphics[width=.49\textwidth]{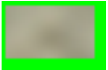}
  \centerline{\scalebox{0.99}{ DRRN~\cite{Tai2017Image}}}
\end{minipage}%
}
\hspace{-0.06in}
\subfigure{
\begin{minipage}[c]{0.19\textwidth}
\centering
  \includegraphics[width=.49\textwidth]{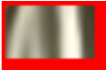}\hspace{-0.03in}
  \includegraphics[width=.49\textwidth]{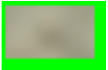}
  \centerline{\scalebox{0.99}{ MemNet~\cite{tai2017memnet}}}
\end{minipage}%
}
\hspace{-0.05in}
\subfigure{
\begin{minipage}[c]{0.19\textwidth}
\centering
  \includegraphics[width=.49\textwidth]{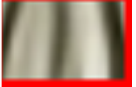}\hspace{-0.03in}
  \includegraphics[width=.49\textwidth]{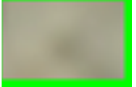}
  \centerline{\scalebox{0.99}{  WaveResNet~\cite{bae2017beyond}}}
\end{minipage}%
}
\hspace{-0.05in}
\subfigure{
\begin{minipage}[c]{0.19\textwidth}
\centering
  \includegraphics[width=.49\textwidth]{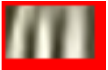}\hspace{-0.03in}
  \includegraphics[width=.49\textwidth]{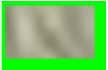}
  \centerline{\scalebox{0.99}{ MWCNN}}
\end{minipage}%
}
\hspace{-0.05in}
\subfigure{
\begin{minipage}[c]{0.19\textwidth}
\centering
  \includegraphics[width=.49\textwidth]{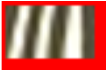}\hspace{-0.03in}
  \includegraphics[width=.49\textwidth]{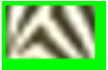}
  \centerline{\scalebox{0.99}{ Ground Truth}}
\end{minipage}%
}
\end{minipage}
\caption{\textit{Single image super-resolution}: result of ``$253027$'' (BSD100) with upscaling factor of $\times$4. }\label{fig:srs4_barbara}
\vspace{-0.03in}
\end{figure*}

\subsubsection{Single image super-resolution}

We also train our MWCNN on SISR task with only the luminance channel, \textit{i.e.} Y in YCbCr color space following \cite{dong2016image}.
$Bicubic$ interpolation is used for image degradation, and upsampling by $bicubic$ interpolation is used before sending degradation image to network.
For qualitative comparisons, we use source codes of nine CNN-based methods, including VDSR~\cite{Kim2015Accurate}, DnCNN~\cite{Zhang2016Beyond}, RED30~\cite{Mao2016Image}, SRResNet~\cite{Ledig2017Photo}, LapSRN~\cite{lai2017deep}, DRRN~\cite{Tai2017Image}, MemNet~\cite{tai2017memnet}, WaveResNet~\cite{bae2017beyond} and SRMDNF~\cite{Zhang_2018_CVPR}.
Since the source code of SRResNet is not released, their results as shown in Table~\ref{tab:sr} are incomplete.
And the results of LapSRN with scale $\times3$ are not listed here since they are not reported in the authors' paper.

Table \ref{tab:sr} summarizes the average PSNR/SSIM results of the competing methods on the four datasets by citing the results in their respective papers.
In terms of both PSNR and SSIM indexes, the proposed MWCNN outperforms other methods in all cases.
Compared with well-known VDSR, our MWCNN achieves a notable gain of about 0.4$\sim$0.8dB by PSNR on Set5 and Set14.
Surprisingly, our MWCNN outperforms VDSR by a large gap with about 1.0$\sim$1.6dB on Urban100.
Even though SRMDNF is trained on RGB space, it is still slightly weaker than our MWCNN.
It can  also  be seen that our MWCNN  outperforms WaveResNet by no less than 0.3dB.
We provide quantitative comparisons with the competing methods on the image ``253027'' from BSD100 in Figure~\ref{fig:srs4_barbara}.
As one can see, our MWCNN can correctly recover the fine and detailed textures, and produce sharp edges due to the frequency and location characteristics of DWT.

\begin{table*}[!htbp]
\centering
\vspace{-0.0in}
\renewcommand\tabcolsep{1.9pt}
\renewcommand\arraystretch{0.99}
\caption{Average PSNR(dB) / SSIM results of the competing methods for JPEG image artifacts removal with quality factors $Q = $ 10, 20, 30 and 40 on datasets Classic5 and LIVE1. Red color indicates the best performance.}
{
\begin{tabular}{ p{1.27cm}<{\centering} p{0.43cm}<{\centering} p{2.12cm}<{\centering} p{2.12cm}<{\centering} p{2.12cm}<{\centering} p{2.12cm}<{\centering} p{2.12cm}<{\centering} p{2.12cm}<{\centering} p{2.12cm}<{\centering} p{1.92cm}<{\centering} }
\hlineB{2.2}
Dataset & $Q$ & JPEG & ARCNN~\cite{Dong2016Compression} & TNRD~\cite{Chen2015Trainable}   & DnCNN~\cite{Zhang2016Beyond}    & MemNet~\cite{tai2017memnet}  &    MWCNN(P) &    MWCNN   \\
\hlineB{2.2}
 \multirow{4}{*}{Classic5}
&10 & 27.82 / 0.7595 & 29.03 / 0.7929 & 29.28 / 0.7992 & 29.40 / 0.8026 &   29.69 / 0.8107 & 30.01 / 0.8195 & \textcolor[rgb]{1,0,0}{30.03} / \textcolor[rgb]{1,0,0}{0.8201} \\
&20 & 30.12 / 0.8344 & 31.15 / 0.8517 & 31.47 / 0.8576 & 31.63 / 0.8610 &   31.90 / 0.8658 & 32.16 / 0.8701 & \textcolor[rgb]{1,0,0}{32.20} / \textcolor[rgb]{1,0,0}{0.8708} \\
&30 & 31.48 / 0.8744 & 32.51 / 0.8806 & 32.78 / 0.8837 & 32.91 / 0.8861 &  - & 33.43 / 0.8930 & \textcolor[rgb]{1,0,0}{33.46} / \textcolor[rgb]{1,0,0}{0.8934}\\
&40 & 32.43 / 0.8911 & 33.34 / 0.8953 &       -        & 33.77 / 0.9003 &  - & 34.27 / 0.9061 & \textcolor[rgb]{1,0,0}{34.31} / \textcolor[rgb]{1,0,0}{0.9063} \\ \hline
\multirow{4}{*}{LIVE1}
&10 & 27.77 / 0.7730 & 28.96 / 0.8076 & 29.15 / 0.8111 & 29.19 / 0.8123 &  29.45 / 0.8193 & 29.69 / 0.8254 & \textcolor[rgb]{1,0,0}{29.70} / \textcolor[rgb]{1,0,0}{0.8260} \\
&20 & 30.07 / 0.8512 & 31.29 / 0.8733 & 31.46 / 0.8769 & 31.59 / 0.8802 &  31.83 / 0.8846 & 32.04 / 0.8885 & \textcolor[rgb]{1,0,0}{32.07} / \textcolor[rgb]{1,0,0}{0.8886} \\
&30 & 31.41 / 0.9000 & 32.67 / 0.9043 & 32.84 / 0.9059 & 32.98 / 0.9090 & - & 33.45 / 0.9153 & \textcolor[rgb]{1,0,0}{33.46} / \textcolor[rgb]{1,0,0}{0.9155} \\
&40 & 32.35 / 0.9173 & 33.63 / 0.9198 &       -        & 33.96 / 0.9247 & - & 34.45 / 0.9301 &  \textcolor[rgb]{1,0,0}{34.47} / \textcolor[rgb]{1,0,0}{0.9300}\\
\hlineB{2.2}
\end{tabular}
}
\label{tab:deblock}
\vspace{-0.06in}
\end{table*}

\begin{figure*}[!htbp]
\setlength{\abovecaptionskip}{0pt}
\setlength{\belowcaptionskip}{0pt}
  \centering

\subfigure{
\begin{minipage}[c]{0.26\textwidth}
\centering
  \includegraphics[width=1\textwidth,height=0.78\textwidth]{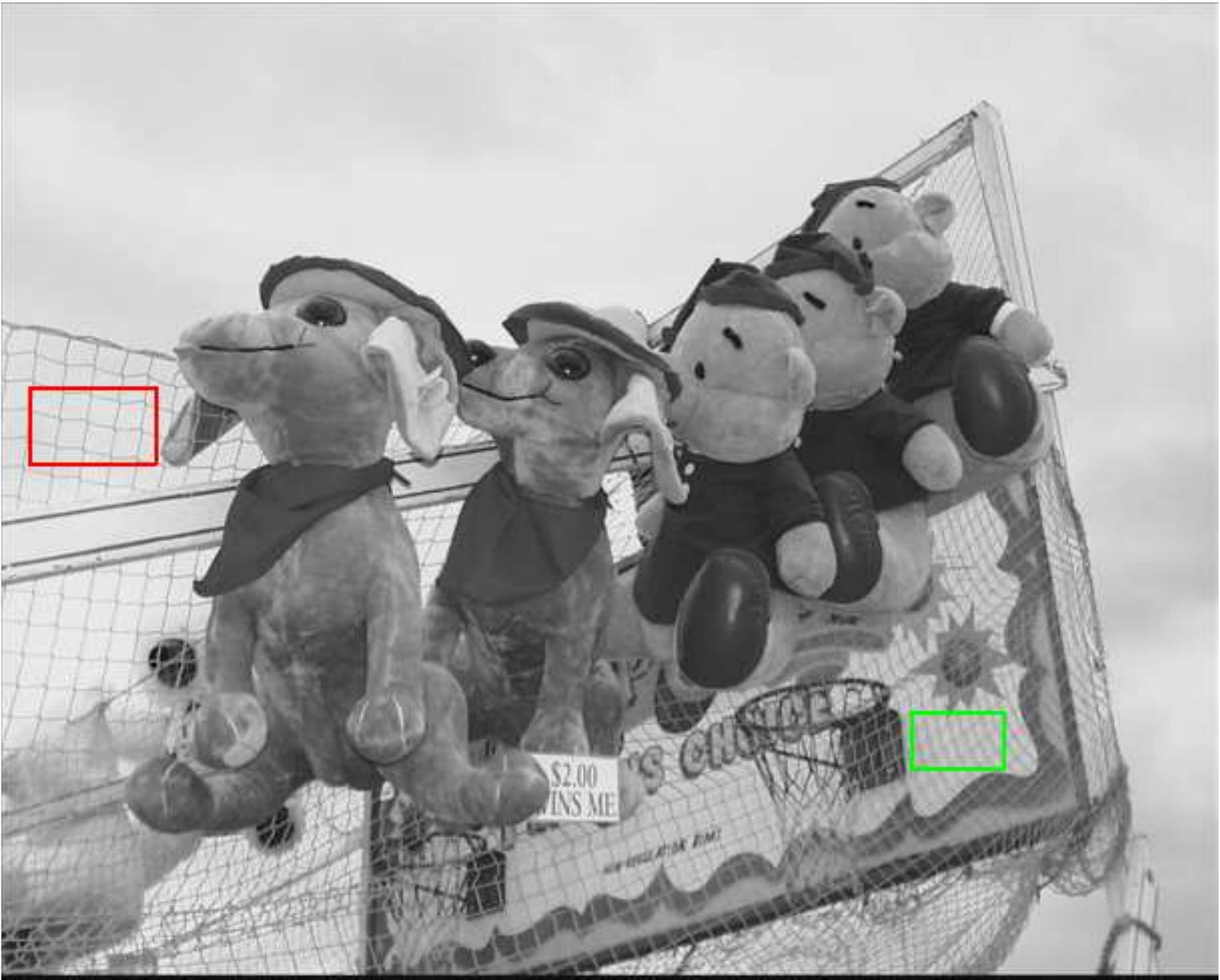}
  \centerline{ Ground Truth}
\end{minipage}%
}
\begin{minipage}[c]{0.71\textwidth}
\vspace{-0.00in}
\subfigure{
\begin{minipage}[c]{0.31\textwidth}
\centering
  \includegraphics[width=.48\textwidth]{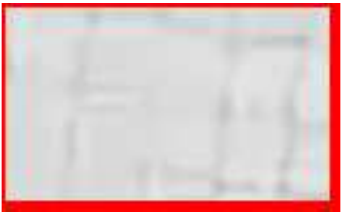}
  \includegraphics[width=.48\textwidth]{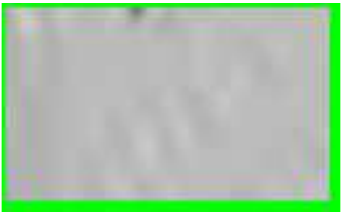}
  \centerline{ ARCNN~\cite{Dong2016Compression} }
\end{minipage}%
}\hspace{0.01in}
\subfigure{
\begin{minipage}[c]{0.31\textwidth}
\centering
  \includegraphics[width=.48\textwidth]{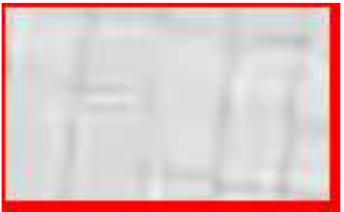}
  \includegraphics[width=.48\textwidth]{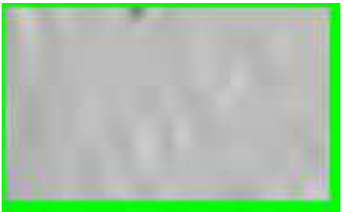}
  \centerline{ TNRD~\cite{Chen2015Trainable} }
\end{minipage}%
}
\vspace{0.00in}
\subfigure{
\begin{minipage}[c]{0.31\textwidth}
\centering
  \includegraphics[width=.48\textwidth]{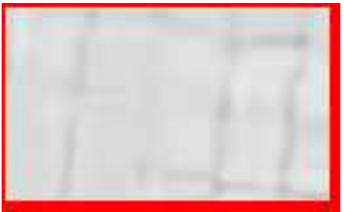}
  \includegraphics[width=.48\textwidth]{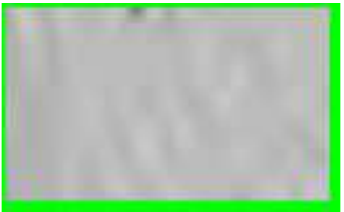}
  \centerline{ DnCNN~\cite{Zhang2016Beyond} }
\end{minipage}%
}\hspace{0.01in}
\vspace{0.1in}

\subfigure{
\begin{minipage}[c]{0.31\textwidth}
\centering
  \includegraphics[width=.48\textwidth]{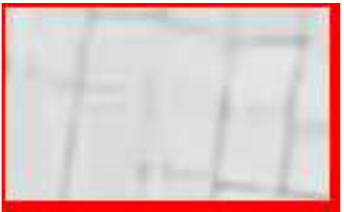}
  \includegraphics[width=.48\textwidth]{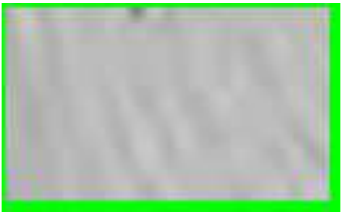}
  \centerline{ MemNet~\cite{tai2017memnet} }
\end{minipage}%
}\hspace{0.01in}
\subfigure{
\begin{minipage}[c]{0.31\textwidth}
\centering
  \includegraphics[width=.48\textwidth]{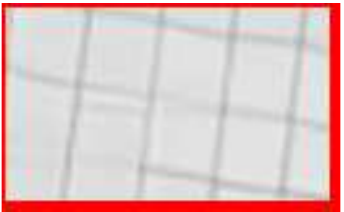}
  \includegraphics[width=.48\textwidth]{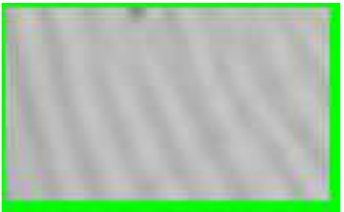}
  \centerline{ MWCNN }
\end{minipage}%
}\hspace{0.01in}
\subfigure{
\begin{minipage}[c]{0.31\textwidth}
\centering
  \includegraphics[width=.48\textwidth]{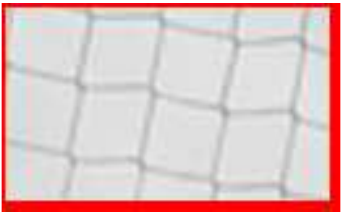}
  \includegraphics[width=.48\textwidth]{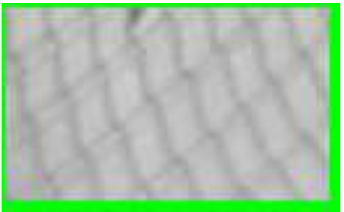}
  \centerline{ Ground Truth}
\end{minipage}%
}
\end{minipage}

\caption{\textit{JPEG image artifacts removal}: visual results of ``$carnivaldolls$'' (LIVE1) with quality factor of 10. }\label{fig:db10_building}
\vspace{-0.01in}
\end{figure*}

\subsubsection{JPEG image artifacts removal}
We apply our method to JPEG image artifacts removal to further demonstrate the applicability of our MWCNN on image restoration.
Here, both JPEG encoder and JPEG image artifacts removal are only focused on the Y channel.
Following \cite{Dong2016Compression}, we consider four settings on quality factor, \textit{e.g.}, $Q$ = 10, 20, 30 and 40, for the JPEG encoder.
In our experiments, MWCNN is compared to four competing methods, \textit{i.e.}, ARCNN~\cite{Dong2016Compression}, TNRD~\cite{Chen2015Trainable}, DnCNN~\cite{Zhang2016Beyond}, and MemNet~\cite{tai2017memnet}.
The results of MemNet~\cite{tai2017memnet} and TNRD~\cite{Chen2015Trainable} are incomplete according to  their paper and released source codes.

Table~\ref{tab:deblock} shows the average PSNR/SSIM results of the competing methods on Classic5 and LIVE1.
Obviously, our  MWCNN obtains superior performance than other methods in terms of quantitative metrics for any of the four quality factors.
Compared to ARCNN on Classic5, our MWCNN surprisingly outperforms by 1dB in terms of  PSNR.
On can see that the PSNR value of MWCNN can be 0.2$\sim$0.3dB higher than those of the second best method (\textit{i.e.}, MemNet~\cite{tai2017memnet}).
In addition to perceptual comparisons, we also provide the image, \textit{i.e.} ``\textit{carnivaldolls}'' form LIVE1  with the quality factor of 10.
Compared with other methods, our  MWCNN is effective in better removing artifacts and restoring detailed textures and sharp salient edges.

\begin{table}[!htbp]\footnotesize
\centering
\caption{Running time (in seconds) of the competing methods for the three tasks on images of size 256$\times$256, 512$\times$512 and 1024$\times$1024:
image denosing is tested on noise level 50, SISR is tested on scale $\times$2, and JPEG image deblocking is tested on quality factor 10.}
\renewcommand\tabcolsep{1.7pt}
\renewcommand\arraystretch{0.99}
\scalebox{0.94}{
\begin{tabular}{ p{1.5cm}<{\centering} p{1.4cm}<{\centering} p{1.4cm}<{\centering} p{1.4cm}<{\centering} p{1.4cm}<{\centering} p{1.4cm}<{\centering} p{1.4cm}<{\centering} }
\hlineB{2.2}
  \multicolumn{6}{ c }{\textbf{Image Denoising}} \\
\hlineB{1.2}
Size & FFDNet~\cite{zhang2018ffdnet} & DnCNN~\cite{Zhang2016Beyond} & RED30~\cite{Mao2016Image} &  MemNet~\cite{tai2017memnet}   &  MWCNN        \\
\hlineB{1.2}
  256$\times$256    & 0.006  &   0.0143   &   1.362    &   0.8775   & 0.0437 \\
  512$\times512$    & 0.012  &   0.0487   &   4.702    &   3.606   & 0.0844 \\
  1024$\times 1024$ & 0.038  &   0.1688   &   15.77    &   14.69   & 0.3343 \\
\hlineB{1.2}
  \multicolumn{6}{ c }{\textbf{Single Image Super-Resolution}} \\
\hlineB{1.2}
Size &  VDSR~\cite{Kim2015Accurate} & LapSRN~\cite{lai2017deep} & DRRN~\cite{Tai2017Image} & MemNet~\cite{tai2017memnet}  &  MWCNN        \\
\hlineB{1.2}
  256$\times$256    & 0.0172 &  0.0229    &    3.063   &   0.8774   & 0.0397 \\
  512$\times512$    & 0.0575 &  0.0357    &    8.050   &   3.605    & 0.0732 \\
  1024$\times 1024$ & 0.2126 &  0.1411    &    25.23   &   14.69    & 0.2876 \\
\hlineB{1.2}
   \multicolumn{6}{ c }{\textbf{JPEG Image Artifacts Removal}} \\
\hlineB{1.2}
Size &   ARCNN~\cite{Dong2016Compression} & TNRD~\cite{Chen2015Trainable} & DnCNN~\cite{Zhang2016Beyond} & MemNet~\cite{tai2017memnet}  &  MWCNN       \\
\bottomrule[0.66pt]
  256$\times$256     &  0.0277  & 0.009   &   0.0157  &   0.8775   & 0.0413 \\
  512$\times512$     &  0.0532  & 0.028   &   0.0568  &   3.607    & 0.0789 \\
  1024$\times 1024$  &  0.1613  & 0.095   &   0.2012  &   14.69    & 0.2717 \\
\hlineB{2.2}
\end{tabular}
}
\label{tab:runtime}
\vspace{-0.01in}
\end{table}

\subsubsection{Running time}
As mentioned previously, the efficiency of CNNs is also an important measure of network performance.
We consider the CNN-based methods with source code and list the GPU running time of the competing methods for the three tasks in Table~\ref{tab:runtime}.
Note that the Nvidia cuDNN-v7.0 deep learning library with CUDA 9.2 is adopted to accelerate the GPU computation under Ubuntu 16.04 system.
In comparison to the state-of-the-art methods, \textit{i.e.}, RED30~\cite{Mao2016Image}, DRRN~\cite{Tai2017Image} and MemNet~\cite{tai2017memnet}, our MWCNN costs far less time but obtain  better performance in terms of PSNR/SSIM metrics.
Meanwhile, our MWCNN is moderately slower by speed but can achieve higher PSNR/SSIM indexes compared  to the other methods.
This means that the effectiveness of MWCNN should be attributed to the incorporation of CNN and DWT rather than  increase of network depth/width.

\subsection{Comparison of MWCNN variants}
Using image denoising and JPEG image artifacts as examples, we mainly focus on two variants of MWCNN:
(i) Ablation experiments to demonstrate where the improved performance comes from.
(ii) The related methods, such as wavelet-based approach and dilated filtering are presented for verifying the effectiveness of the proposed method.
Note that MWCNN with 24-layer is employed as our baseline, and all the MWCNN variants are designed using the same architecture for fair comparison.

\begin{table}[!htbp]\footnotesize
\centering
\renewcommand\tabcolsep{1.7pt}
\renewcommand\arraystretch{0.99}
\caption{
Performance comparison of ablation experiment in terms of average PSNR (dB) and running time (in seconds): image denosing is tested on noise level 50 and JPEG image deblocking is tested on quality factor 10.
}
\scalebox{0.601}{

\begin{tabular}{ p{1.05cm}<{\centering} p{1.55cm}<{\centering} p{1.75cm}<{\centering} p{1.65cm}<{\centering} p{1.9cm}<{\centering} p{1.95cm}<{\centering} p{1.95cm}<{\centering} p{1.85cm}<{\centering} p{1.55cm}<{\centering} }
\toprule[1pt]
   \multicolumn{7}{ c }{\textbf{Image Denoising ($\sigma=50$)}} \\
\bottomrule[0.66pt]
Dataset &   U-Net~\cite{Ronneberger2015U} & U-Net~\cite{Ronneberger2015U}+S  & U-Net~\cite{Ronneberger2015U}+D & MWCNN (P+C) & MWCNN (Haar) &   MWCNN (DB2)  &   MWCNN (HD)      \\ \hline 
Set12    & 27.42 / 0.079 &  27.41 / \textbf{0.074}  & 27.46 / 0.080 & 27.76 / 0.081 & 27.79 / 0.075   & \textbf{27.81} / 0.127 &  27.77 / 0.091 \\
Set68    & 26.30 / 0.076 &  26.29 / \textbf{0.071} & 26.21 / 0.075 & 26.54 / 0.077 & 26.58 / 0.072   & \textbf{26.59} / 0.114 &  26.57 / 0.086  \\
Urban100 & 26.68 / 0.357 &  26.72 / \textbf{0.341}  & 26.99 / 0.355 & 27.46 / 0.354 & 27.53 / 0.346   & \textbf{27.55} / 0.576 &  27.50 / 0.413  \\
\bottomrule[0.66pt]
   \multicolumn{7}{ c }{\textbf{JPEG Image Artifacts Removal ($Q=10$)}} \\
\bottomrule[0.66pt]
Classic5   & 29.61 / 0.093  & 29.60 / \textbf{0.082} & 29.68 / 0.097 & 30.02 / 0.091 &  30.03 / 0.083 & \textbf{30.04} / 0.185 &  29.99 / 0.115 \\
LIVE1      & 29.36 / 0.112  & 29.36 / \textbf{0.109} & 29.43 / 0.120 & 29.69 / 0.120 &  29.70 / 0.111  & \textbf{29.71} / 0.234 &  29.68 / 0.171 \\
\bottomrule[1pt]
\end{tabular}
}
\label{tab:denoising_part}
\vspace{-0.01in}
\end{table}

\subsubsection{Ablation experiments}
Ablation experiments are provided for verifying the effectiveness of additionally embedded wavelet:
(i) the default U-Net with the same architecture to MWCNN,
(ii) U-Net+S: using sum connection instead of concatenation,
and (iii) U-Net+D: adopting learnable conventional downsampling filters, \textit{i.e.} convolution operation with stride 2 to replace max pooling.
We also compare with the modified MWCNN(P) method which adds one layer of CNN right after inputting and another layer of CNN before outputting, and denote it as MWCNN(P+C).
Three MWCNN variants with different wavelet transform are also considered, including:
(i) MWCNN (Haar): the default MWCNN with Haar wavelet,
(ii) MWCNN (DB2): MWCNN with \textit{Daubechies-2} wavelet,
and (iii) MWCNN (HD): MWCNN with Haar in contracting subnetwork and \textit{Daubechies-2} in expanding subnetwork.

Table~\ref{tab:denoising_part} lists the PSNR and running time results of these methods.
We have the following observations.
(i) The ablation experiments indicate that adopting sum connection instead of concatenation can slightly improve efficiency with almost no decrease of PNSR.
(ii) Due to the biorthogonal and time-frequency localization properties of wavelet, our wavelet based method possesses more powerful abilities for image restoration.
The pooling operation causes the loss of high-frequency information  and leads to  difficulty of recovering damaged image.
Our MWCNN can easily outperform U-Net+D method which adopts learnable downsampling filters.
This indicates that learning alone is not enough and the violation of the invertibility can cause information loss.
(iii) Compared with MWCNN (P+C), the proposed method still performs slightly better despite the fact that MWCNN (P+C) has more layers, thereby verifying the effectiveness of the proposed method.
(iv) Compared with MWCNN (DB2) and MWCNN (HD), using Haar wavelet for downsampling and upsampling in network is the best choice in terms of quantitative evaluation.
MWCNN (Haar) has similar running time as dilated CNN and U-Net but achieves higher PSNR results, which demonstrates the effectiveness of MWCNN for trading off between performance and efficiency.

\begin{table}[!htbp]\footnotesize

\centering
\renewcommand\tabcolsep{1.7pt}
\renewcommand\arraystretch{0.99}
\caption{
Performance comparison of MWCNN variants in terms of average PSNR (dB) and running time (in seconds): image denosing is tested on noise level 50 and JPEG image deblocking is tested on quality factor 10.
}
\scalebox{0.705}{

\begin{tabular}{ p{1.19cm}<{\centering} p{1.55cm}<{\centering} p{1.55cm}<{\centering} p{1.55cm}<{\centering} p{1.55cm}<{\centering} p{1.75cm}<{\centering} p{1.95cm}<{\centering}}
\toprule[1pt]
   \multicolumn{7}{ c }{\textbf{Image Denoising ($\sigma=50$)}} \\
\bottomrule[0.66pt]
Dataset &      Dilated~\cite{yu2015multi} &      Dilated-2  & DCF~\cite{Han2017Framing} & DCF+R & WaveResNet~\cite{bae2017beyond} &  MWCNN (Haar)     \\
\bottomrule[0.66pt]
Set12    & 27.45 / 0.181 & 24.81 / 0.185 &   27.38 /  0.081 & 27.61 / 0.081 & 27.49 / 0.179  & \textbf{27.79} / \textbf{0.075}   \\
Set68    & 26.35 / 0.142 & 24.32 / 0.174 &   26.30 /  0.075 & 26.43 / 0.075 & 26.38 / 0.143  & \textbf{26.58} / \textbf{0.072}    \\
Urban100 & 26.56 / 0.764 & 24.18 / 0.960 &   26.65 / 0.354  & 27.18 / 0.354 & - / -  & \textbf{27.53} / \textbf{0.346}    \\
\bottomrule[0.66pt]
   \multicolumn{7}{ c }{\textbf{JPEG Image Artifacts Removal ($Q=10$)}} \\
   \bottomrule[0.66pt]
Classic5   & 29.72 / 0.287 &  29.49 / 0.302 & 29.57 / 0.104 & 29.88 / 0.104  & - / -  &   \textbf{30.03} / \textbf{0.083}  \\
LIVE1      & 29.49 / 0.354 &  29.26 / 0.376 & 29.38 / 0.155 & 29.63 / 0.155  & - / -  &  \textbf{29.70} / \textbf{0.111}  \\
\bottomrule[1pt]
\end{tabular}
}
\label{tab:variants}
\vspace{-0.00in}
\end{table}

\subsubsection{More MWCNN variants}
We compare the PSNR results by using more related MWCNN variants.
Two 24-layer dilated CNNs are provided: (i) Dilated: the hybrid dilated convolution~\cite{Wang2017Understanding} to suppress the gridding effect, and
(ii) Dilated-2: the dilate factor of all layers is set to 2 following the gridding effect.
The WaveResNet method in~\cite{bae2017beyond} is provided for comparison.
Moreover, since the source code of deep convolutional framelets (DCF) is not available, a re-implementation of deep convolutional framelets (DCF) without residual learning~\cite{Ye2017Deep} is also considered in the experiments.
DCF with residual learning (denoted as DCF+R) is provided for fair comparison.

Table~\ref{tab:variants} lists the PSNR and running time results of these methods.
We have the following observations.
(i) The gridding effect with the sparse sampling and inconsistence of local information authentically has adverse influence on restoration performance.
(ii) The worse performance of DCF also indicates that independent processing of subbands harms  intra-frequency information dependency.
(iii) The fact that our MWCNN method sightly outperforms the DCF+R method means that the adverse influence of independent processing can be eliminated after several non-linear operations.

\subsubsection{Hierarchical level of MWCNN}
Here, we discuss the suitable level of our MWCNN even if it can be extended to higher level.
Nevertheless, deeper network and heavier computational burden also come with higher level.
Thus, we select a suitable level for better balance between efficiency and performance.
As shown in Table~\ref{tab:depth}, the PSNR and running time results of MWCNNs with the levels of 0 to 4 (\textit{i.e.}, MWCNN-0 $\sim$ MWCNN-4) are reported.
We note that MWCNN-0 is a 6-layer CNN without WPT.
In terms of the PSNR metric, MWCNN-3 is much better than MWCNN-1 and MWCNN-2, while negligibly weaker than  MWCNN-4.
Meanwhile, the speed of MWCNN-3 is also moderate compared with other levels.
Based on the above analysis, we choose MWCNN-3 as the default setting.

\begin{table}[!htbp]\footnotesize
\renewcommand\tabcolsep{1.2pt}
\renewcommand\arraystretch{0.99}
\centering
\caption{Average PSNR (dB) and running time (in seconds) of MWCNNs with different levels on Gaussian denoising with the noise level of 50.}
\scalebox{0.9}{
\begin{tabular}{ p{1.3cm}<{\centering} p{1.6cm}<{\centering} p{1.6cm}<{\centering} p{1.6cm}<{\centering} p{1.6cm}<{\centering} p{1.6cm}<{\centering}}
\toprule[1pt]
Dataset & MWCNN-0 &  MWCNN-1 & MWCNN-2 & MWCNN-3 & MWCNN-4           \\
\bottomrule[0.66pt]
Set12      & 26.84 / 0.017 & 27.25 / 0.041   &    27.64 / 0.064    &    27.79 / 0.073     &    27.80 / 0.087  \\
Set68      & 25.71 / 0.016 & 26.21 / 0.039   &    26.47 / 0.060    &    26.58 / 0.070     &    26.59 / 0.081  \\
Urban100   & 25.98 / 0.087 & 26.53 / 0.207   &    27.12 / 0.298    &    27.53 / 0.313     &    27.55 / 0.334  \\
\bottomrule[0.99pt]
\end{tabular}
}
\label{tab:depth}
\vspace{-0.02in}
\end{table}

\subsection{Extend to Object Classification}


%
Using object classification as an example, we test our MWCNN on six famous benchmarks: CIFAR-10, CIFAR-100~\cite{krizhevsky2009learning}, SVHN~\cite{netzer2011reading}, MNIST~\cite{lecun1998gradient}, ImageNet1K~\cite{krizhevsky2012imagenet} and Places365~\cite{zhou2017places} for our evaluation.
CIFAR-10 and CIFAR-100 consist of 60000 32$\times$32 colour images in 10 classes, including 50,000 images for training and 10,000 images for testing.
MNIST~\cite{lecun1998gradient} is handwritten digit database with 28$\times$28 resolution, which has a training set of 60,000 examples, and a test set of 10,000 examples.
The SVHN dataset contains more than  600,000 digit images.
ImageNet1K is a resized dataset which consists of two resolutions,  32$\times$32 denoted as ImageNet32, and 64$\times$64 as ImageNet64.
Places365 is resized to $100\times100$ for training and testing in our work.
Here, we modify and compare several CNN methods, such as pre-activation ResNet (PreResNet)~\cite{he2016identity}, All-CNN~\cite{springenberg2014striving}, WideResNet~\cite{zagoruyko2016wide}, PyramidNet~\cite{han2017deep}, DenseNet~\cite{huang2017densely} and ResNet~\cite{he2016deep}.
Specifically, we follow \cite{wang2018global} to verify our MWCNN on ResNet architecture.
As our MWCNN, we use DWT transformation with $1\times 1$ convolution to instead of avg-pooling operation as described in Sec.~\ref{sec:classarch}, and denote it as `MW', while the original CNN is denoted as `Base'.

Table~\ref{tab:classifacation1} shows the detailed results of accuracy with the competing methods PreResNet~\cite{he2016identity}, All-CNN~\cite{springenberg2014striving}, WideResNet~\cite{zagoruyko2016wide}, PyramidNet~\cite{han2017deep}, DenseNet~\cite{huang2017densely} on CIFAR-10, CIFAR-100, SVHN, MNIST and ImageNet32.
Table~\ref{tab:classifacation2} and Table~\ref{tab:classifacation3} show Top-1 and Top-5 error of ResNet~\cite{he2016deep} on imagenet64 and Place365.
One can see that MWCNN can easily surpass the original CNN because of the powerful DWT used.
Our MWCNN is quite different from DCF~\cite{Ye2017Deep}:
DCF combines CNN with DWT during decomposition, where different CNNs are deployed for each subband.
However, the results in Table~\ref{tab:denoising_part} indicates that independent processing of subbands is not suitable for image restoration.
In contrast, MWCNN incorporates DWT into CNN from the perspective of enlarging receptive field without information loss, allowing embedding DWT with any CNNs with pooling operations.
By taking all subbands as input, MWCNN is more powerful in modeling inter-band dependency.
Moreover, our MWCNN is formulated as a single, generic, plug-and-play module that can be used as a direct replacement of downsampling operation without any adjustments in
the network architecture.

\begin{table}[!htbp]\footnotesize
\centering
\renewcommand\tabcolsep{1.4pt}
\renewcommand\arraystretch{0.99}
\caption{Accuracy on CIFAR-10, CIFAR-100, SVHN, MNIST and ImageNet32.}
\scalebox{0.78}{
\begin{tabular}{ p{3.0cm}<{\centering} p{1.1cm}<{\centering} p{1.2cm}<{\centering} p{1.35cm}<{\centering} p{1.0cm}<{\centering} p{1.0cm}<{\centering} p{1.3cm}<{\centering} }
\toprule[1pt]
Model &Dataset & CIFAR-10 &  CIFAR-100 & SVHN  & MNIST & ImageNet32  \\
\bottomrule[1pt]
\multirow{2}{*}{DenseNet-BC-100(k=12)~\cite{huang2017densely}}
&Base    &   95.40    &  77.38   &   98.03  & 99.69 & 48.32 \\
&MW    &     \bf{95.57}   &     \bf{77.70}   &  \bf{98.11}  & \bf{99.74} & \bf{49.77} \\ \hline
\multirow{2}{*}{AllConv~\cite{springenberg2014striving}}
&Base    &   91.58    &  67.57   & 98.06  & 99.67 &  30.12 \\
&MW    &      \bf{93.08}  &     \bf{71.06}   & \bf{98.09}  & \bf{99.70} & \bf{36.64} \\ \hline
\multirow{2}{*}{PyramidNet-164~\cite{han2017deep}}
&Base    &   96.09     &    80.35  &  97.98  & 99.72 & 55.30 \\
&MW    &   \bf{96.11}   &   \bf{80.49}    &   \bf{98.17}  & \bf{99.74} & \bf{55.51} \\ \hline
\multirow{2}{*}{PreResNet-164($\alpha=28$)~\cite{he2016identity}}
&Base    &  95.29     &  77.32  & 98.04  & 99.67 & 46.16  \\
&MW    &    \bf{95.72}    &   \bf{77.67}     &   \bf{98.09}  & \bf{99.72} & \bf{47.88} \\ \hline
\multirow{2}{*}{WideResNet-28-10~\cite{zagoruyko2016wide}}
&Base    &    96.56   &  81.30   & 98.19   & 99.70 & 57.51 \\
&MW    &   \bf{96.60}    &     \bf{81.42}   & \bf{98.36}  & \bf{99.75} & \bf{57.66} \\
\bottomrule[1pt]
\end{tabular}
}
\label{tab:classifacation1}
\vspace{-0.1in}
\end{table}

\begin{table}[!htbp]\footnotesize
\centering
\renewcommand\tabcolsep{1.7pt}
\renewcommand\arraystretch{0.99}
\caption{Top-1 and Top-5 error on ImageNet64.}
{
\begin{tabular}{ p{2.5cm}<{\centering} p{1.5cm}<{\centering} p{1.9cm}<{\centering} p{1.9cm}<{\centering}}
\toprule[1pt]
Model &ImageNet64 & Top-1 & Top-5  \\
\bottomrule[1pt]
\multirow{2}{*}{ResNet18-512d~\cite{he2016deep}}
&Base      &  49.08     &    24.25     \\
&MW        &  \bf{48.46}      &     \bf{23.96}      \\ \hline
\multirow{2}{*}{ResNet50~\cite{he2016deep}}
&Base    &   43.28    &  19.39   \\
&MW    &      \bf{41.70}  &     \bf{18.05}     \\ \hline

\multirow{2}{*}{ResNet50-512d~\cite{he2016deep}}
&Base    &   41.42    &  18.14     \\
&MW    &     \bf{41.27}   &     \bf{17.70}  \\
\bottomrule[1pt]
\end{tabular}
}
\label{tab:classifacation2}
\vspace{-0.1in}
\end{table}

\begin{table}[!htbp]\footnotesize
\centering
\renewcommand\tabcolsep{1.7pt}
\renewcommand\arraystretch{0.99}
\caption{Top-1 and Top-5 error on Place365.}
{
\begin{tabular}{ p{2.5cm}<{\centering} p{1.5cm}<{\centering} p{1.9cm}<{\centering} p{1.9cm}<{\centering} }
\toprule[1pt]
Model &Dataset & Top-1 & Top-5  \\
\bottomrule[1pt]
\multirow{2}{*}{ResNet18-512d~\cite{he2016deep}}
&Base      &  49.96     &    19.19     \\
&MW        &  \bf{49.56}      &     \bf{18.88}      \\
\bottomrule[1pt]
\end{tabular}
}
\label{tab:classifacation3}
\vspace{-0.01in}
\end{table}

\section{Conculsion}\label{sec:con}
In this paper, we present MWCNN to better trade off the receptive field and efficiency.
To this end, DWT is introduced as a downsampling operation to reduce spatial resolution and enlarge the receptive field, and can be embedded into any CNNs using pooling operation.
More specifically, MWCNN takes both low-frequency and high-frequency subbands as input and is safe to perform downsampling without information loss.
In addition, WPT can be treated as pre-defined parameters to ease network learning.
We first design an architecture for image restoration based on U-Net, which consists of a contracting sub-network and a expanding subnetwork (for object classification, only the contracting network is employed).
Due to the invertibility of DWT and its frequency and location property, the proposed MWCNN is effective in recovering detailed textures and sharp structures from degraded observation.
Extensive experiments demonstrate the effectiveness and efficiency of MWCNN on three restoration tasks, \textit{i.e.}, image denoising, SISR, and JPEG image artifacts removal, and object classification task when using different CNN methods.
In future work, we aims at designing novel network architecture to extend MWCNN to more restoration tasks such as image deblurring.
We will also investigate flexible MWCNN models for handling blind restoration tasks.
High-level dense prediction tasks, such as object detection and image segmentation, often is accomplished by adopting pooling for downsampling and then performing upsampling for dense prediction. Therefore, the limitations of pooling operation may still be unfavorable in these tasks.
In the future work, we will modify to extend the proposed MWCNN to these tasks for better preserving of fine-scale features.

\section*{Acknowledgment}

This work was supported in part by the National Natural Scientific Foundation of China (NSFC) under Grant No. 61872118, 61773002 and 61671182.

\end{document}